%% file: main.tex
\documentclass[10pt,twocolumn,letterpaper]{article}

\usepackage{cvpr}              % To produce the CAMERA-READY version
\input{preamble}

\definecolor{cvprblue}{rgb}{0.21,0.49,0.74}
\usepackage[pagebackref,breaklinks,colorlinks,allcolors=cvprblue]{hyperref}
\usepackage{multirow} 
\usepackage{array}
\usepackage{colortbl}
\usepackage[ruled,vlined]{algorithm2e}
\usepackage[accsupp]{axessibility}  % Improves PDF readability for those with disabilities.

\definecolor{LightGray}{RGB}{240,240,240}

\newcolumntype{L}[1]{>{\raggedright\let\newline\\\arraybackslash\hspace{0pt}}m{#1}}
\newcolumntype{C}[1]{>{\centering\let\newline\\\arraybackslash\hspace{0pt}}m{#1}}
\newcolumntype{R}[1]{>{\raggedleft\let\newline\\\arraybackslash\hspace{0pt}}m{#1}}

\usepackage{pifont}

\definecolor{checkmark}{HTML}{008000}
\definecolor{cross}{HTML}{FF0000}

\def\paperID{32467} % *** new
\def\confName{CVPR}
\def\confYear{2026}

\title{MultiModalPFN: Extending Prior-Data Fitted Networks \\ for Multimodal Tabular Learning}

\author{
  Wall Kim\\
  Samsung Electronics\\
  Hwaseong, South Korea\\
  {\tt\small wall.kim@samsung.com}
  \and
  Chaeyoung Song \qquad Hanul Kim\\
  Seoul National University of Science and Technology\\
  Seoul, South Korea\\
  {\tt\small \{cysong, hukim\}@seoultech.ac.kr} 
}

\begin{document}
\maketitle
\input{sec/0_abstract}    
\input{sec/1_intro}
\input{sec/2_related}
\input{sec/3_method}
\input{sec/4_experiment}
\input{sec/5_conclusion}

\section*{Acknowledgements}
This work was supported in part by the National Research Foundation of Korea (NRF) funded by the Korean government under Grants RS-2023-00221365 and RS-2024-00352566.

{
    \small
    \bibliographystyle{ieeenat_fullname}
    \bibliography{main}
}

\input{sec/X_suppl}

\end{document}

%% file: preamble.tex
 \usepackage{url}

%% file: sec/0_abstract.tex
\begin{abstract}
Recently, TabPFN has gained attention as a foundation model for tabular data. However, it struggles to integrate heterogeneous modalities such as images and text, which are common in domains like healthcare and marketing, thereby limiting its applicability. To address this, we present the Multi-Modal Prior-data Fitted Network (MMPFN), which extends TabPFN to handle tabular and non-tabular modalities in a unified manner. MMPFN comprises per-modality encoders, modality projectors, and pre-trained foundation models. The modality projectors serve as the critical bridge, transforming non-tabular embeddings into tabular-compatible tokens for unified processing. To this end, we introduce a multi-head gated MLP and a cross-attention pooler that extract richer context from non-tabular inputs while mitigates attention imbalance issue in multimodal learning. Extensive experiments on medical and general-purpose multimodal datasets demonstrate that MMPFN consistently outperforms competitive state-of-the-art methods and effectively exploits non-tabular modalities alongside tabular features. These results highlight the promise of extending prior-data fitted networks to the multimodal setting, offering a scalable and effective framework for heterogeneous data learning. The source code is available at \url{https://github.com/too-z/MultiModalPFN}.

\end{abstract}

%% file: sec/1_intro.tex
\section{Introduction}\label{sec:intro}
Tabular data is one of the most widely used data formats across domains such as healthcare, finance, and marketing. Traditionally, gradient-boosted decision trees~\cite{chen2016xgboost, ke2017lightgbm, prokhorenkova2018catboost} have dominated this field, owing to their fast training and strong predictive performance. However, recent progress in modern tabular deep learning models~\cite{somepalli2021saint, bahri2022scarf} has shown that deep architectures can learn more expressive tabular representations and often surpass traditional tree-based methods.

These advances have also broadened the scope of tabular data analysis to multimodal settings, where structured features are combined with unstructured modalities such as images and text~\cite{hager2023best}; for example, diagnostic tasks may jointly leverage structured test results and medical images~\cite{huang2020fusion, schilcher2024fusion}, while marketing applications may integrate numerical sales records with textual product reviews~\cite{das2024towards, sukel2024multimodal}. Despite this growing interest, attempts to extend gradient-boosted decision trees to heterogeneous data types have yielded only modest gains, and deep learning models that jointly embed tabular data with images or text, while promising, often suffer from limited performance in data-scarce regimes and slow training~\cite{cui2023deep, zhao2024deep}.

More recently, TabPFN~\cite{hollmann2023tabpfn, hollmann2025accurate} has attracted considerable attention as a tabular foundation model that treats supervised learning on tables as amortized Bayesian inference, achieving strong performance on small- and medium-sized datasets in a single forward pass. While TabPFN establishes a powerful prior over purely tabular distributions, its pretraining is restricted to synthetic tabular data, and no principled extensions to unstructured modalities have been explored. Consequently, despite its strong performance on purely tabular tasks, TabPFN does not address the growing need to jointly model tabular features with image and text modalities in practical multimodal applications.

In this work, we propose the Multi-Modal Prior-data Fitted Network (MMPFN), an extension of TabPFN that processes tabular and non-tabular modalities in a unified manner. MMPFN first extracts features with per-modality encoders for tabular, image, and text inputs. A modality projector then aligns the non-tabular embeddings with the tabular embedding space. The resulting multimodal embeddings are fed into the pretrained TabPFN backbone, so that its tabular prior can be directly reused while incorporating information from images and text through light fine-tuning. In addition, MMPFN explicitly tackles two common failure modes in multimodal learners: overcompressed non-tabular embeddings and attention imbalance from token-count disparities, by introducing a multi-head gated MLP (MGM) that expands non-tabular representations into multiple tokens and a cross-attention pooler (CAP) that compresses them into a compact, balanced set. MGM and CAP constitute our modality projector. We evaluate MMPFN on multiple benchmarks~\cite{pacheco2020pad, sawyer2016curated, petfinder_adoption_2019, airbnb, salary, cloth} that pair tabular inputs with images or text inputs. Across nearly all datasets, MMPFN surpasses recent state-of-the-art methods~\cite{hager2023best, du2024tip, hemker2024healnet, luo2025time, bonnier2024revisiting, tang2024bag}. Extensive experiments demonstrate that MGM and CAP effectively mitigate the identified failure modes, while MMPFN scales positively as modalities are added, preserves the strengths of TabPFN’s modeling in low-data regimes. Our main contributions are summarized as follows:
\begin{itemize}
    \item We propose MMPFN, the first framework to extend TabPFN, pretrained on synthetic tabular distributions, to heterogeneous inputs (tabular + image/text) through a unified pathway.
    \vspace{0.1cm}
    \item We identify two failure modes: overcompressed non-tabular embeddings and token-count–induced attention imbalance, and introduce MGM and CAP as components of the modality projector to address them.
    \vspace{0.1cm}
    \item Through experiments on medical and general-purpose datasets, we show that MMPFN outperforms competitive baselines, scales positively as modalities are added, and maintains robust performance under data scarcity and limited compute.
\end{itemize}

%% file: sec/2_related.tex
\section{Related works}\label{sec:related}
\paragraph{Vision--Language Multimodal Models.}
Early research in multimodal learning developed fusion and conditioning mechanisms for integrating text and images. FiLM~\cite{perez2018film} introduced feature-wise modulation for language-conditioned visual reasoning, while early transformer-based models such as ViLBERT, VisualBERT, VL-BERT, LXMERT, and UNITER~\cite{lu2019vilbert,li2019visualbert,su2019vl,tan2019lxmert,chen2020uniter} explored co-attention and unified architectures, achieving state-of-the-art results on vision--language benchmarks. A major shift came with CLIP~\cite{radford2021learning}, which used large-scale contrastive pretraining for scalable zero-shot transfer. More recent approaches, such as BLIP-2~\cite{li2023blip} and LLaVA~\cite{liu2023visual}, integrated large language models for generalizable multimodal reasoning.

\vspace*{-0.3cm}
\paragraph{Tabular and Multimodal Models.}
The pretraining-driven paradigm has since expanded to structured data. In the tabular domain, approaches typically adopt either a \textit{row-as-text} strategy, serializing entire rows for large language model (LLM) processing~\cite{hegselmann2023tabllm}, or a \textit{per-column embedding} strategy with modality-specific encoders. Methods such as Tab2Text~\cite{lin2024tab2text} transform rows into textual narratives for improved alignment, while others~\cite{bonnier2024revisiting} demonstrate that careful design of fusion layers substantially improves benchmarks. LANISTR~\cite{ebrahimi2023lanistr} extended this direction with similarity-based multimodal masking, enabling joint learning from language, images, and structured inputs even with missing modalities.

Unstructured--structured integration has also been explored in image-centric datasets. Representative works included MMCL~\cite{hager2023best}, which aligned tabular and image embeddings through contrastive learning; TIP~\cite{du2024tip}, which improved robustness to missing features; STiL~\cite{du2025stil}, which leveraged unlabeled data through semi-supervised pseudo-labeling; TIME~\cite{luo2025time}, which used TabPFN~\cite{hollmann2023tabpfn} as a tabular encoder; and Turbo~\cite{jiang2025multimodal}, which strengthened cross-modal reasoning. Beyond individual models, toolkits such as AutoGluon~\cite{tang2024autogluon} and modular pipelines~\cite{gu-budhkar-2021-package} provided practical infrastructure for multimodal integration. Despite this progress, most work remained focused on vision--language tasks, and systematic treatment of structured data remained limited. Fusion strategies were often heuristic and less reliable under low-data regimes or modality imbalance. These gaps motivated more general multimodal tabular systems.

\vspace*{-0.3cm}
\paragraph{General-Purpose Pre-trained Models.}
Pretraining large foundation models has transformed representation learning across domains. In NLP, models progressed from masked language modeling to more efficient self-supervised strategies such as ELECTRA~\cite{clark2020electra} and DeBERTa~\cite{he2021deberta}. Later refinements including DeBERTaV3~\cite{he2023debertav3}, ModernBERT~\cite{modernbert}, and multilingual encoders such as BGE/M3~\cite{bge_m3} introduced architectural improvements (e.g., disentangled embeddings, FlashAttention-2, optimized tokenization) and broadened applications to retrieval and cross-lingual tasks.

In computer vision, self-supervised pretraining has become a dominant paradigm. DINOv2~\cite{oquabdinov2} and DINOv3~\cite{simeoni2025dinov3} showed scalable self-distillation for robust visual features, EVA~\cite{EVA,eva02} advanced masked image modeling with large Vision Transformers, and iBOT~\cite{zhou2021ibot} combined masking and self-distillation for effective ViT representations. For structured data, TabPFN~\cite{hollmann2023tabpfn,hollmann2025accurate} extended this idea by pretraining on large synthetic datasets to learn a general prior over tabular distributions. It achieved strong performance on small and medium-sized datasets in a single forward pass without fine-tuning, making it a foundation model for tabular learning. Yet pretraining for multimodal tabular data remained underexplored relative to NLP and vision. Bridging this gap is important for multimodal foundation models with structured inputs.

%% file: sec/3_method.tex
\begin{figure*}[!t]
\centering
\includegraphics[width=\linewidth]{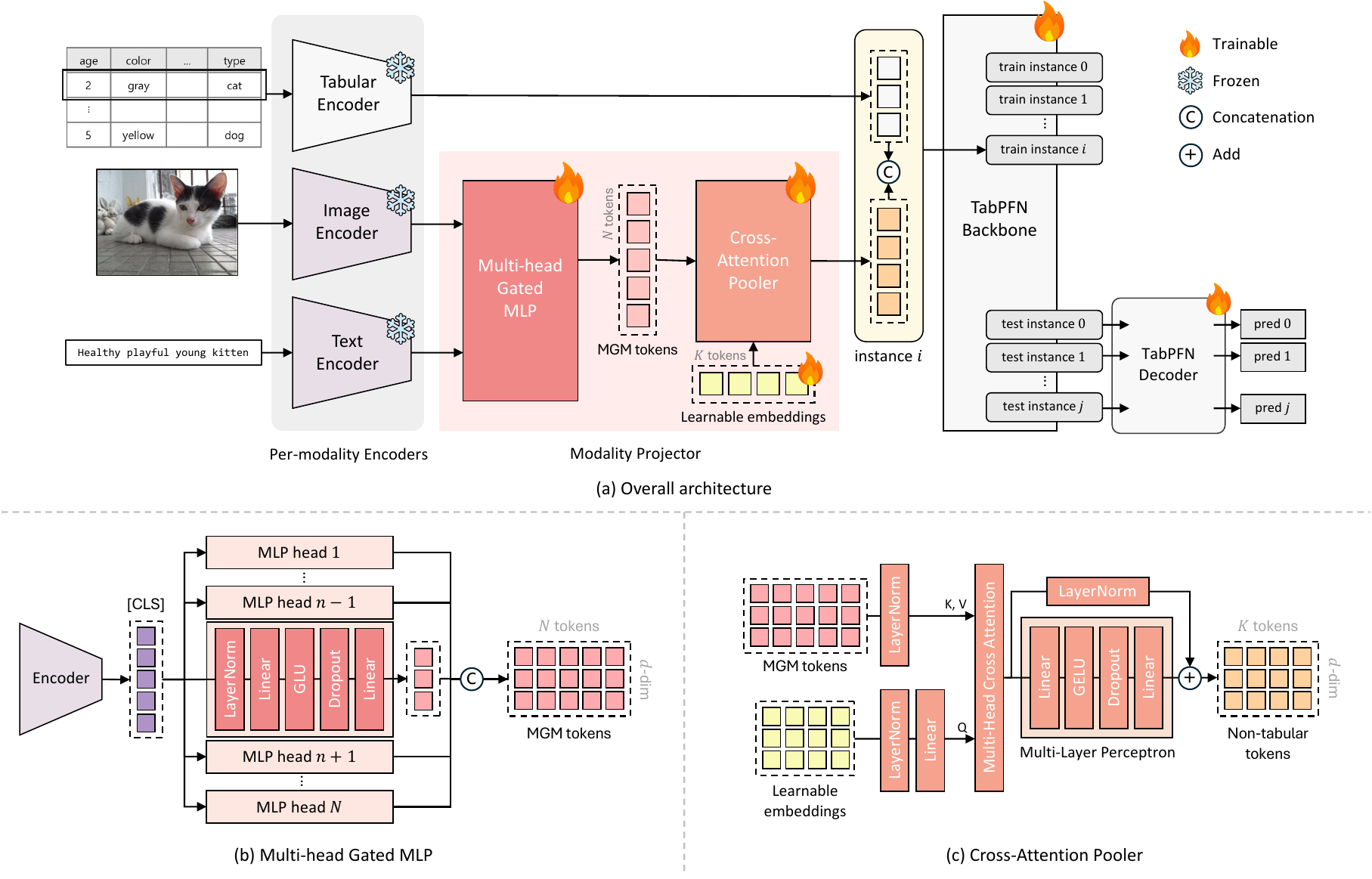}
\caption{\textbf{An overview of MMPFN.} MMPFN extends TabPFN by incorporating per-modality encoders and a modality projector to extract features from non-tabular data. Newly developed components are highlighted in color, while existing ones appear in gray. Layers marked as ‘frozen’ remain fixed during fine-tuning, whereas all others are trainable. Encoded target labels are part of the training inputs but are omitted from the diagram for clarity.\label{fig:architecture}}
\end{figure*}

\section{Proposed Method} \label{sec:method}
\Cref{fig:architecture}~(a) illustrates the overall architecture of our multimodal PFN (MMPFN) that extends TabPFN~\cite{hollmann2023tabpfn, hollmann2025accurate} to the multimodal setting, where image or text modalities accompany tabular inputs. Therefore, we begin by briefly reviewing TabPFN. We then describe the proposed multimodal PFN architecture, including the per-modality encoders and the modality projector that aligns non-tabular embeddings with the tabular feature space, followed by the training protocol for fine-tuning MMPFN on downstream multimodal tasks. Finally, we analyze the phenomenon of attention imbalance that arises from token-count disparities across modalities.

\subsection{Preliminary: TabPFN}\label{subsec:method_preliminary}
TabPFN is a tabular foundation model that treats tabular learning as amortized Bayesian inference. More specifically, a transformer is pretrained on a large collection of synthetic tabular datasets sampled from structural causal model priors. During pretraining, it learns to map a small labeled training set and an accompanying query set directly to posterior-predictive label distributions in a single forward pass. Once pretrained, TabPFN can be applied to new tabular tasks without task-specific optimization, simply by feeding the new training–test pairs to the network.

Architecturally, TabPFN stacks 2D TabPFN blocks. Each block splits attention into two stages: feature attention, where each feature attends to other features within the same sample, and sample attention, where the same feature attends across all samples. This design yields permutation invariance over both samples and features and scales efficiently to larger tables than those encountered during pretraining. For in-context inference, TabPFN processes the concatenated training and test rows with masks that allow self-attention within labeled training rows and restrict test rows to cross-attend only to training rows. An MLP head then maps the test embeddings to the predictions. 

\subsection{Multimodal PFN: Architecture}\label{subsec:method_architecture}
As shown in~\Cref{fig:architecture}~(a), MMPFN consists of per-modality encoders, a modality projector, and a TabPFN backbone. The per-modality encoders map each input modality to a feature representation, while the modality projector aligns image and text embeddings with the shared tabular embedding space. The TabPFN backbone then jointly processes the resulting multimodal embeddings, and a lightweight decoder head produces predictions for the test samples.

\paragraph{Per-Modality Encoders.}
The per-modality encoders comprise tabular, image, and text branches. The tabular branch is identical to the TabPFN v2 encoder~\cite{hollmann2025accurate} and remains frozen during fine-tuning. For images, we employ the DINOv2 ViT-B/14 backbone~\cite{oquabdinov2,dosovitskiy2021image}: input images are resized so that both height and width are divisible by 14, and the final \texttt{[CLS]} token is used as a global image representation. For text, we adopt an ELECTRA-based encoder~\cite{clark2020electra}, chosen based on preliminary experiments in which it consistently outperformed DeBERTa variants~\cite{he2021deberta}. Text inputs are tokenized and truncated to a maximum length of 512 tokens, and the corresponding \texttt{[CLS]} embedding is used as the text representation.

\paragraph{Modality Projector.}
The modality projector transforms image and text embeddings into tabular-like representations, which share $d$-dimensional space compatible with the TabPFN backbone. It comprises two sublayers: a multi-head gated MLP (MGM) and a cross-attention pooler (CAP). MGM addresses the limitation of a single \texttt{[CLS]} embedding, which can overly compress image/text information, by expanding it into $N$ parallel $d$-dimensional projections. \Cref{fig:architecture}~(b) illustrates the detailed structure of MGM. Specifically, the \texttt{[CLS]} embedding is fed into $N$ MLP heads that project the encoder output dimension to $d$ and produce candidate modality-specific tokens. A Gated Linear Unit (GLU)~\cite{dauphin2017language} modulates the contribution of each head, encouraging head-wise specialization and preserving diverse aspects of the original non-tabular representation in the resulting token set.

CAP then balances tabular and non-tabular cues before fusion in the TabPFN backbone. As shown in~\Cref{fig:architecture}~(c), it takes the $N$ MGM tokens as keys and values and introduces $K$ learnable query vectors that cross-attend to them, yielding $K$ representative $d$-dimensional embeddings per modality. The pooled tokens are refined by an MLP. These $K$ tokens form a compact, calibrated summary of image/text information and are concatenated with the tabular tokens along the feature dimension to construct the multimodal input table for TabPFN. Without pooling, too many non-tabular tokens can induce attention imbalance, where the modality with more tokens dominates the attention budget and suppresses tabular signal. CAP mitigates this by producing a compact, calibrated set of embeddings for the TabPFN backbone. We analyze this in~\Cref{sec:method_analysis}.

\subsection{Multimodal PFN: Training}\label{sec:method_trainig}
Since TabPFN is pre-trained on large corpora of synthetic tabular data, its representations can be misaligned with image/text embeddings. We therefore freeze all modality encoders and train the modality projector, the TabPFN backbone, and the decoder. Note that all components are pre-trained, except for the modality projector. To leverage TabPFN’s in-context inference, we follow its standard protocol: split the multimodal data into training and test sets, concatenate their embeddings into a single table, and feed it to the backbone. The model then produces predictions for the test samples to obtain supervisory signals for training.

\subsection{Attention Imbalance in MMPFN} \label{sec:method_analysis}
We study how the number of non-tabular tokens affects attention mechanism. Consider a query token $q$ attending to two sets of keys: non-tabular tokens $k^{(I)}_{1},\cdots,k^{(I)}_{N_I}$ and tabular tokens $k^{(T)}_{1},\cdots,k^{(T)}_{N_T}$, where $N_I$ and $N_T$ are their respective counts. The scaled dot-product attention scores are given by
\begin{equation}
s^{(I)}_i = q^\top k^{(I)}_i/{\sqrt d}, \qquad
s^{(T)}_j = q^\top k^{(T)}_j/{\sqrt d}.
\end{equation}
Let $w^{(I)}_i = e^{s^{(I)}_i}$ and $w^{(T)}_j = e^{s^{(T)}_j}$ be the unnormalized attention weights, and define the per-token expectations $c_I = \mathbb{E}[w^{(I)}_i]$ and $c_T = \mathbb{E}[w^{(T)}_j]$, where the expectation is over token indices and any randomness in $(q,k)$. Also, let $a_I$ denote the total attention weight allocated to the non-tabular set, defined by 
\begin{equation}
a_I = \sum_{i=1}^{N_I}
\frac{w^{(I)}_i}{\sum_{u=1}^{N_I} w^{(I)}_u + \sum_{v=1}^{N_T} w^{(T)}_v}.
\end{equation}
Then its expectation is approximated by
\begin{equation}
\mathbb{E}[a_I] \approx \frac{N_I c_I}{N_I c_I + N_T c_T}
\label{eq:expected_weight}
\end{equation}
Hence, when per-token quality is comparable ($c_I \approx c_T$), token-count imbalance ($N_I > N_T$) induces attention imbalance, potentially degrading performance. Consequently, MMPFN’s performance might vary with the modality token ratio. This suggests the importance of CAP. In~\Cref{sec:experiments}, we validate this observation by varying $K$ in the CAP.

%% file: sec/4_experiment.tex
\renewcommand{\arraystretch}{1.2} 

\begin{table*}[!t]
\scriptsize
\centering
\caption{\textbf{Statistics of the multimodal datasets used in our experiments.} 
“\# images” and “\# text” denote the number of image and text fields per sample, respectively.
PetFinder variants indicate which modalities (tabular (T), image (I), text (t)) are used.\label{tab:dataset}}
\vspace*{-0.1cm}
{\setlength{\tabcolsep}{6.5pt}
\begin{tabular}{l|cccccccc}
\toprule
\textbf{Dataset} & \textbf{\# train samples} & \textbf{\# test samples} & \textbf{\# features} & \textbf{\# numeric features} & \textbf{\# categorical features} & \# \textbf{images} & \textbf{\# text} & \textbf{\# classes} \\
\midrule
PAD-UFES-20~\cite{pacheco2020pad} & 1838 & 460 & 21 & 3 & 18 & 1 & 0 & 6 \\
CBIS-DDSM(Mass)~\cite{sawyer2016curated} & 1318 & 378 & 8 & 3 & 5 & 3 & 0 & 2 \\
CBIS-DDSM(Calc)~\cite{sawyer2016curated} & 1545 & 326 & 8 & 3 & 5 & 3 & 0 & 2 \\
Airbnb~\cite{airbnb} & 18316 & 4579 & 50 & 27 & 23 & 0 & 1 & 10 \\
Salary~\cite{salary} & 15841 & 3961 & 4 & 1 & 3 & 0 & 3 & 6 \\
Cloth~\cite{cloth} & 18788 & 4698 & 5 & 2 & 3 & 0 & 3 & 5 \\
PetFinder-I (T+I)~\cite{petfinder_adoption_2019} & 11721 & 2931 & 19 & 5 & 14 & 1 & 0 & 5 \\
PetFinder-t (T+t)~\cite{petfinder_adoption_2019} & 11721 & 2931 & 19 & 5 & 14 & 0 & 1 & 5 \\
PetFinder-A (T+I+t)~\cite{petfinder_adoption_2019} & 11721 & 2931 & 19 & 5 & 14 & 1 & 1 & 5 \\
\bottomrule
\end{tabular}}
\end{table*}

\section{Experiments} \label{sec:experiments}

\subsection{Experimental Setup} 
\paragraph{Dataset.}
We evaluate MMPFN using well-established multimodal datasets that have been extensively validated in previous studies~\cite{mraz2025towards, tang2024bag, tang2024autogluon, bonnier2024revisiting, jiang2024tabular}:
\begin{itemize}
    \item PAD-UFES-20(PU20)~\cite{pacheco2020pad} contains 2{,}298 samples from six skin lesion types, each paired with a clinical image and up to 26 metadata features. 
    \item CBIS-DDSM~\cite{sawyer2016curated} is a curated subset of DDSM with digitized mammograms, annotated regions of interest for calcifications and masses, biopsy-verified benign/malignant labels, and lesion-level metadata.
    \item Airbnb~\cite{airbnb} provides a detailed snapshot of Melbourne’s homestay activity as of December 2018; following the preprocessing strategy of TTT, we discretize the target into ten quantile-based groups of equal size.
    \item Salary~\cite{salary} contains 15{,}841 training and 3{,}961 test job postings from India, each with company, years of experience, job description, designation, job type, key skills, and location, with the task of predicting the salary range.
    \item Cloth~\cite{cloth} comprises 23{,}486 customer reviews with text fields and 10 tabular features for multimodal sentiment and recommendation prediction.
    \item PetFinder~\cite{petfinder_adoption_2019}, released for the Kaggle PetFinder.my challenge, contains over 14{,}000 pet profiles with images, descriptive text, and structured attributes for predicting a five-class adoption-speed outcome.
\end{itemize}
For each dataset, we randomly split the data into training and test sets, except for CBIS-DDSM, for which we use the predefined train-test split. \Cref{tab:dataset} summarizes the statistics of these datasets. More details about datasets are provided in the supplementary material.

\vspace*{-0.15cm}
\paragraph{Implementation Details.}
We fine-tune MMPFN for $100$ iterations using cross-entropy loss. We use AdamW~\cite{loshchilov2019decoupled} with a learning rate of $1 \times 10^{-5}$ and batch size $1$. For all experiments, we use random seeds $\{0,1,2,3,4\}$ and report average accuracy. More details are provided in the supplementary material.

\footnotetext{We cite all results of \citet{luo2025time} directly. Although TIME used the CBIS-DDSM dataset without specifying subtype, the reported sample size matches the calcification subset, so we list it under CBIS-DDSM calcification in Table~\ref{table:result_image}. Since the code is unavailable, reproduction was infeasible.}

\subsection{Main Results}
\paragraph{Results on Tabular–Image Modality Datasets.}
\Cref{table:result_image} summarizes the classification accuracy of MMPFN and state-of-the-art baselines on four tabular–image datasets. Compared with fine-tuned TabPFN~\cite{hollmann2025accurate}, which uses only tabular inputs, MMPFN consistently improves performance by leveraging image features. MMCL~\cite{hager2023best}, TIP~\cite{du2024tip}, and HEALNet~\cite{hemker2024healnet} show inconsistent results, likely due to the small dataset size and low-dimensional tabular features. In contrast, MMPFN achieves the best results on all datasets except Mass, where it remains competitive with the top models. Compared with TIME~\cite{luo2025time}, which uses TabPFN as its tabular encoder, MMPFN delivers substantial gains, suggesting that our modality projection strategies are more effective than simple fusion. CatBoost~\cite{prokhorenkova2018catboost} uses image embeddings as raw input features and achieves strong performance. AutoGluon~\cite{tang2024autogluon}, an AutoML framework for multimodal data, also performs competitively on several benchmarks. However, MMPFN achieves a better average rank than both models.

\begin{table}[!t]
\scriptsize
\centering
\caption{\textbf{Comparison with state-of-the-art on tabular-image multimodal datasets}. Results are reported as accuracy (rank), averaged over five random seeds, where lower rank indicates better performance. ``Avg.'' denotes the mean accuracy across datasets. Best and second-best results are marked in \textbf{bold} and \underline{underline}.}
\vspace*{-0.1cm}
\setlength{\tabcolsep}{5.8pt}
\begin{tabular}{l|cccc|c}
\toprule
\textbf{Method} & \textbf{PU20} & \textbf{Mass} & \textbf{Calc} & \textbf{Petfinder} & \textbf{Avg.} \\ 
\midrule
TabPFN~\cite{hollmann2023tabpfn}  & \underline{82.17 (2)} & 71.27 (5) & \underline{73.31 (2)} & 36.33 (8) &  4.25 \\
\midrule
Catboost~\cite{prokhorenkova2018catboost} & 80.43 (4) & \textbf{78.31 (1)} & 72.09 (4) & 38.69 (4) & \underline{3.25}\\
AutoGluon~\cite{tang2024autogluon} & 81.09 (3) & \underline{76.28 (2)} & 71.04 (6) & 38.81 (3) &  3.50\\
MMCL~\cite{hager2023best} & 76.61 (7) & 57.62 (7) & 60.12 (8) & 36.61 (7) & 7.25\\
TIP~\cite{du2024tip} & 78.75 (6) & 73.12 (4) & 67.96 (7) & 37.28 (5) & 5.50\\
HEALNet~\cite{hemker2024healnet} & 74.65 (8) & 68.10 (6) & 71.83 (5) & 37.03 (6) & 6.25\\
TIME~\cite{luo2025time} & 80.35 (5) & - & 72.70 (3)\footnotemark & \underline{39.25 (2)} & 3.33\\
\midrule
\rowcolor{gray!10}
MMPFN & \textbf{85.22 (1)} & 74.53 (3) & \textbf{75.40 (1)} & \textbf{40.74 (1)} & \textbf{1.50}\\
\bottomrule
\end{tabular}
\label{table:result_image}
\end{table}

\vspace*{-0.2cm}
\paragraph{Results on Tabular–Text Modality Datasets.}
\Cref{table:result_text} reports results on tabular–text datasets. As in the image setting, adding text features consistently improves over the fine-tuned TabPFN baseline. MMPFN is particularly strong on Airbnb, which includes 50+ tabular features and a single text field, allowing tabular-specialized models to capture most of the predictive signal. Accordingly, MMPFN substantially outperforms language model--based methods, such as TFN~\cite{zadeh2017tensor} and MulT~\cite{tsai2019multimodal}, which struggle to exploit abundant tabular features. By contrast, Cloth has few informative tabular features, while the review text carries most of the signal. This is reflected in the weak performance of tabular-only models and the strong results of AllTextBERT~\cite{bonnier2024revisiting}, indicating that text-specialized models excel in such cases. Even so, among methods that explicitly preserve tabular structure, MMPFN achieves the best overall performance, trailing the text-specialized baseline by only a small margin. This contrasts with prior multimodal tabular studies, which focused on tabular-dominant datasets~\cite{hager2023best}. Overall, MMPFN effectively handles both tabular and unstructured modalities.

\begin{table}[!t]
\centering
\caption{\textbf{Comparison with state-of-the-art on tabular-text multimodal datasets}. Results are reported as accuracy (rank), averaged over five random seeds, where lower rank indicates better performance. ``Avg.'' denotes the mean accuracy across datasets. Best and second-best results are marked in \textbf{bold} and \underline{underline}.}
\vspace*{-0.1cm}
\setlength{\tabcolsep}{5.8pt}
\scriptsize
\begin{tabular}{l|cccc|c}
\toprule
 \textbf{Method} & \textbf{Airbnb} & \textbf{Salary} & \textbf{Cloth} & \textbf{Petfinder} & \textbf{Avg.}\\ 
\midrule
TabPFN~\cite{hollmann2023tabpfn} & \underline{46.96 (2)} & 44.96 (6) & 55.07 (9) & 36.33 (7) & 6.00\\
\midrule
Catboost~\cite{prokhorenkova2018catboost} & 43.56 (4) & 40.36 (9) & 59.24 (8) & 35.47 (8) & 7.25\\
AutoGluon~\cite{tang2024autogluon} & 44.60 (3) & 45.24 (5) & \textbf{72.07 (1)} & 37.96 (4) & \underline{3.25}\\
AllTextBERT~\cite{bonnier2024revisiting} & 30.9 (9) & 44.0 (7) & 68.0 (3) & 34.6 (9) & 7.00\\
TFN~\cite{zadeh2017tensor} & 35.7 (8) & 45.8 (3) & 60.1 (7) & 36.8 (6) & 6.00\\
MulT~\cite{tsai2019multimodal} & 36.3 (7) & 45.4 (4) & 63.6 (6) & 37.6 (5) & 5.50\\
TTT~\cite{bonnier2024revisiting} & 38.3 (6) & \textbf{47.2 (1)} & 65.5 (5) & 38.9 (3) & 3.75\\
TabSTAR~\cite{arazi2025tabstar} & 40.06 (5) & 43.75 (8) & \underline{71.75 (2)} & \textbf{41.53 (1)} & 4.00\\
\midrule
\rowcolor{gray!10}
MMPFN & \textbf{47.78 (1)} & \underline{46.17 (2)} & 66.26 (4) & \underline{39.04 (2)} & \textbf{2.25}\\
\bottomrule
\end{tabular}
\label{table:result_text}
\end{table}

\footnotetext{AllTextBert converts all tabular features into strings, concatenates them, and inputs the resulting sequence into \texttt{DistilBERT-base-uncased} for modeling, as described in~\cite{bonnier2024revisiting}.}

\subsection{Analysis}
\begin{figure}[!t]
\centering
\includegraphics[width=\linewidth]{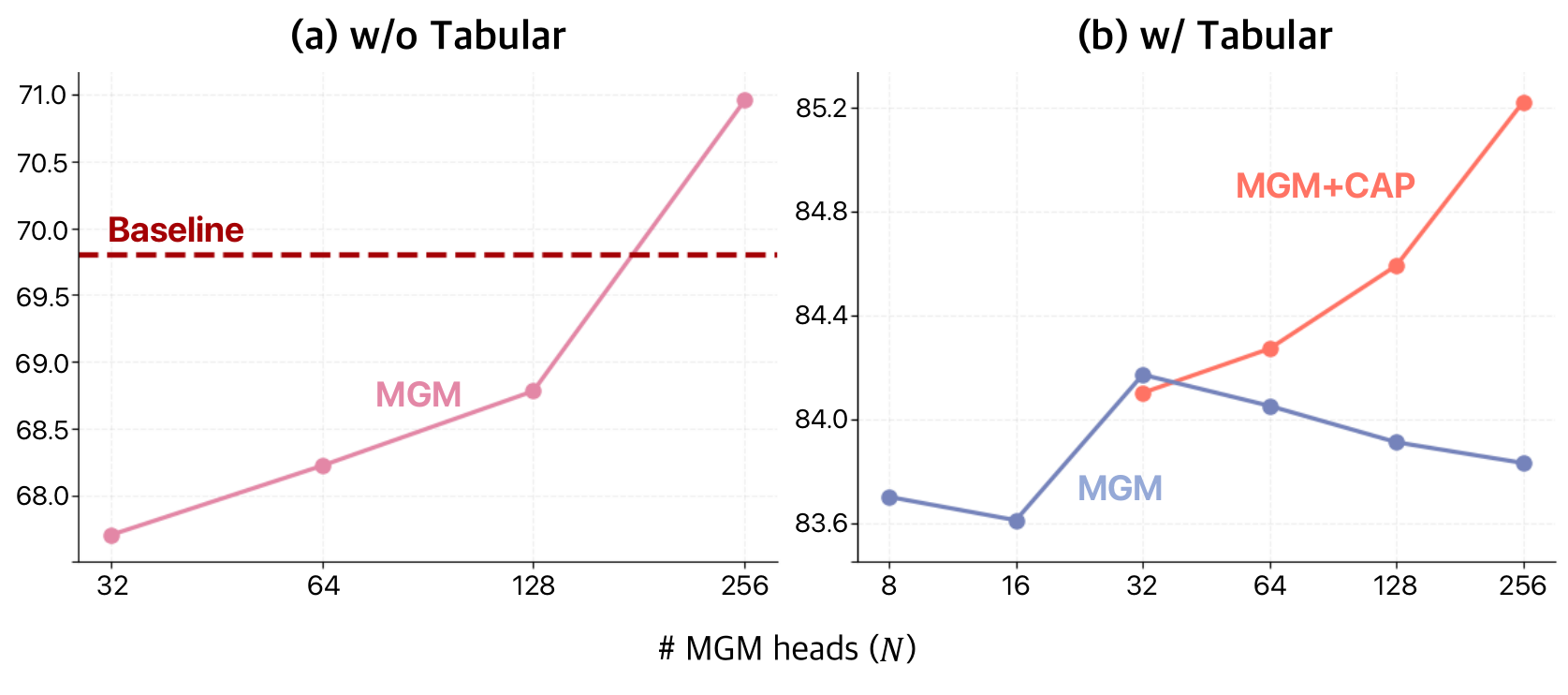}
\caption{\textbf{Performance on PU20 versus the number of non-tabular tokens.} (a) Image-only results with DINOv2 and an MLP baseline. (b) Multimodal results under token imbalance. The y-axis shows accuracy and the x-axis shows the number of MGM heads. In (b), MGM+CAP uses 24 CAP heads.}
\label{fig:perf_by_n_nontab_pu20}
\end{figure}

\paragraph{MMPFN as an Image and Text Classifier.}
\Cref{fig:perf_by_n_nontab_pu20}~(a) evaluates MMPFN with non-tabular–only inputs. As a baseline, we use pre-trained DINOv2~\cite{oquabdinov2} or Electra~\cite{clark2020electra} $[CLS]$ embeddings with an attached MLP classifier, a widely used and often near-optimal setup~\cite{sun2019fine}. MMPFN uses only MGM to generate non-tabular embeddings from the CLS token. With these embeddings as the sole inputs, MMPFN remains within $\sim$1\% of DINOv2 (69.30\% vs.\ 69.89\%). Accuracy increases with the number of image tokens, suggesting that additional tokens capture complementary, higher-resolution information. Although trained on synthetic tabular data, MMPFN effectively classifies image embeddings mapped to tabular-like features, showing that it is not limited to native tabular inputs. CAP further provides a modest gain.

\paragraph{Attention Imbalance.}
We empirically analyze attention imbalance in multimodal processing. In~\Cref{fig:perf_by_n_nontab_pu20}~(a), MMPFN uses only non-tabular inputs. Accuracy increases with the number of MGM heads, showing that additional non-tabular tokens improve performance. CAP is not used here. These results show that a PFN trained on synthetic tabular data can classify projected non-tabular features competitively with standard baselines. TabPFN is therefore not limited to native tabular inputs.

\begin{figure}[!t]
\centering
\includegraphics[width=\linewidth]{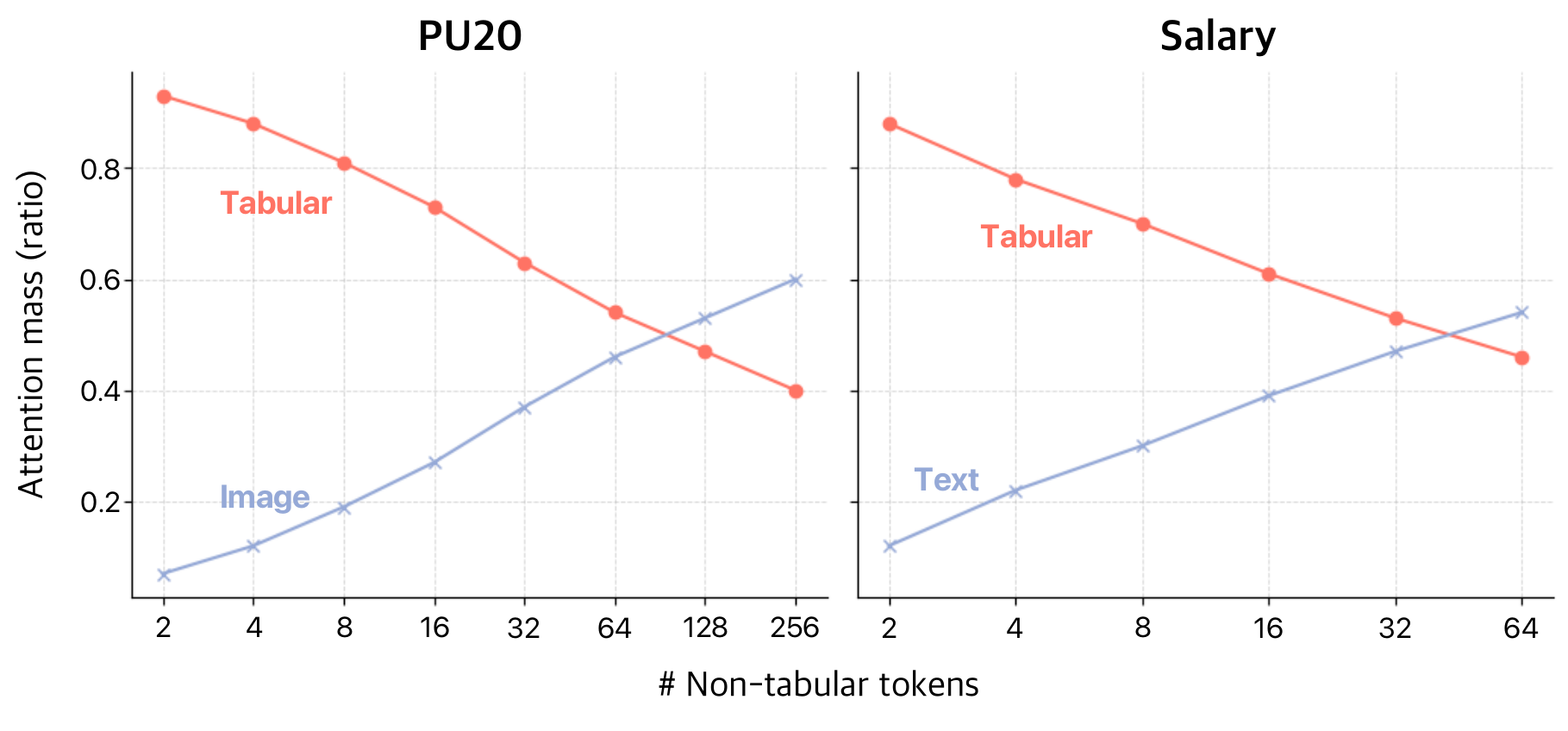}
\caption{\textbf{Token count and attention mass}. Attention mass for tabular and non-tabular tokens is measured as the number of non-tabular tokens varies, without CAP. Values are averaged over 12 self-attention layers in TabPFN. PU20 and Salary use 11 and 4 tabular tokens. The x-axis is in log scale.}
\label{fig:token_attention}
\end{figure}

In~\Cref{fig:perf_by_n_nontab_pu20}~(b), the trend changes when tabular and non-tabular features are used together. MMPFN performs best when the two modalities have similar token counts, and performance drops as one modality dominates the sequence. This pattern does not appear in~\Cref{fig:perf_by_n_nontab_pu20}~(a), where only non-tabular inputs are used. The contrast is consistent with attention imbalance: the modality with more tokens absorbs more of the attention budget, while the other receives less attention and contributes less signal. MGM+CAP mitigates this problem by extracting non-tabular features with enough MGM heads and then compressing them into 24 CAP tokens. As the number of MGM heads increases, MGM+CAP improves steadily and outperforms MGM alone.

We further examine attention allocation by varying the number of non-tabular tokens while keeping the number of tabular tokens fixed. In~\Cref{fig:token_attention}, attention mass shifts monotonically toward the non-tabular modality as its token count increases, while the mass on tabular tokens decreases. This result supports the same explanation: a modality with more tokens receives a larger share of the attention budget. \Cref{tab:token_representation} then separates the effects of token count and representation quality. Increasing token count alone reduces MGM by nearly 2\%p despite identical representations, whereas MGM+CAP remains stable. When representation quality improves at the same total token count ($4\times32$ and 128), MGM still suffers from the larger token set, but MGM+CAP benefits from the stronger representations by compressing them into a compact set of tokens. More details are provided in the supplementary material.

\begin{table}[!t]
\centering
\scriptsize
\caption{\textbf{Token count and representation quality on PU20 and Cloth}. We compare non-tabular token configurations on PU20 (left) and Cloth (right). Columns \textbf{32} and \textbf{128} denote $N{=}32$ and $N{=}128$ MGM heads, and \textbf{4$\times$32} denotes four repetitions of the $N{=}32$ setting. For MGM+CAP, $K$ is fixed to 24 and 4, respectively.}
\setlength{\tabcolsep}{3.7pt}
\begin{tabular}{l|ccc}
\toprule
Method & \textbf{32} & \textbf{4$\times$32} & \textbf{128} \\
\midrule
MGM     & 84.17 & 82.43 & 83.91 \\
MGM+CAP & 84.10 & 83.57 & 84.59 \\
\bottomrule
\end{tabular}
\hspace{0.03cm}
\begin{tabular}{l|ccc}
\toprule
Method & \textbf{32} & \textbf{4$\times$32} & \textbf{128} \\
\midrule 
MGM     & 63.12 & 61.76 & 61.86 \\
MGM+CAP & 64.20 & 64.22 & 65.03  \\
\bottomrule
\end{tabular}
\label{tab:token_representation}
\end{table}

\paragraph{Modality Projector.}
\Cref{tab:ablation_glu} ablates the modality projector, comparing single-head (Linear, MLP) and multi-head (MLP, MoE, MGM) variants. Parameter budgets are provided in the supplementary material. Single-head baselines apply one linear or MLP projection to the non-tabular \texttt{[CLS]} embedding. Although they improve over the tabular-only backbone, they yield the lowest average accuracies, suggesting that a single projection overcompresses non-tabular information. Expanding \texttt{[CLS]} into multiple tokens consistently improves performance, showing the benefit of capturing diverse aspects of image/text features. Within this family, MoE is less effective and less stable across datasets, suggesting that sparse expert routing is hard to exploit in the low-data regime. By contrast, MGM combines multi-head projection with GLU-based gating and achieves the best accuracy on every dataset and the best overall average.

We next examine how the modality projector incorporates non-tabular information. Specifically, we compare CAP, which pools non-tabular features into representative tokens, with Feature-wise Linear Modulation (FiLM)~\cite{perez2018film}, which uses non-tabular representations to generate feature-wise affine parameters for tabular tokens. As shown in~\Cref{tab:ablation_cap}, CAP consistently outperforms FiLM on all datasets. This result suggests that controlling token count is important for tabular--non-tabular fusion because it alleviates attention imbalance between the two modalities. FiLM applies the same channel-wise transformation to all tabular tokens, limiting token-specific use of non-tabular cues. CAP instead compresses non-tabular features into representative tokens that the TabPFN encoder attends jointly with tabular tokens, preserving relevant cross-modal cues while reducing token imbalance.

\begin{table*}[!t]
\centering
\caption{\textbf{Ablation of the design of MGM.} We compare single-head and multi-head feature extraction mechanisms (Linear/MLP, MoE, and the proposed MGM) across all datasets. MoE denotes Mixture-of-Experts. Best results are highlighted in \textbf{bold}.}
\scriptsize
{\setlength{\tabcolsep}{9.2pt}
\begin{tabular}{l|l|ccccccccc|c}
\toprule
Category & Method & PU20 & Mass & Calc & Cloth & Salary & Airbnb & PetFinder-I & PetFinder-T & PetFinder-A & Avg. \\
\midrule
\multirow{2}{*}{Single-head}
& Linear         & 83.48 & 66.19 & 74.17 & 58.06 & 44.67 & 47.02 & 37.22 & 36.64 & 37.30 & 53.86 \\
& MLP & 83.78 & 64.87 & 73.87 & 60.44 & 44.33 & 46.85 & 37.22 & 36.64 & 37.30 & 54.14 \\
\midrule
\multirow{3}{*}{Multi-head}
& MLP & 84.39 & 67.99 & 73.99 & 64.39 & 45.93 & 46.79 & 40.64 & 38.12 & 40.04 & 55.81 \\
& MoE            & 83.22 & 66.67 & 73.13 & 55.07 & 44.25 & 46.12 & 36.82 & 36.83 & 36.94 & 53.23 \\
\cmidrule(lr){2-12}
& MGM            & \textbf{85.22} & \textbf{74.53} & \textbf{75.40} & \textbf{66.26} & \textbf{46.17} & \textbf{47.78} & \textbf{40.70} & \textbf{39.04} & \textbf{41.19} & \textbf{57.37} \\
\bottomrule
\end{tabular}}
\label{tab:ablation_glu}
\vspace{0.1cm}
\end{table*}

\begin{table*}[!t]
\centering
\scriptsize
\caption{\textbf{Ablation of the design of CAP.} We compare two mechanisms for incorporating non-tabular information (FiLM and the proposed CAP) on all datasets. FiLM denotes Feature-wise Linear Modulation. Best results are highlighted in \textbf{bold}.}
\setlength{\tabcolsep}{11pt}
\begin{tabular}{l|ccccccccc|c}
\toprule
Method & PU20 & Mass & Calc & Cloth & Salary & Airbnb & PetFinder-I & PetFinder-T & PetFinder-A & Avg. \\
\midrule
FiLM~\cite{perez2018film} & 80.00 & 73.02 & 73.25 & 65.82 & 42.27 & 46.06 & 39.84 & 38.31  & 40.77 & 55.48    \\
CAP  & \textbf{85.22} & \textbf{74.53} & \textbf{75.40} & \textbf{66.26} & \textbf{46.17} & \textbf{47.78} & \textbf{40.70} & \textbf{39.04} & \textbf{41.19} & \textbf{57.37}\\
\bottomrule
\end{tabular}
\label{tab:ablation_cap}
\end{table*}

\begin{figure}[!t]
\centering

% Row 1
\begin{minipage}{0.48\columnwidth}
  \centering
  \includegraphics[width=\linewidth]{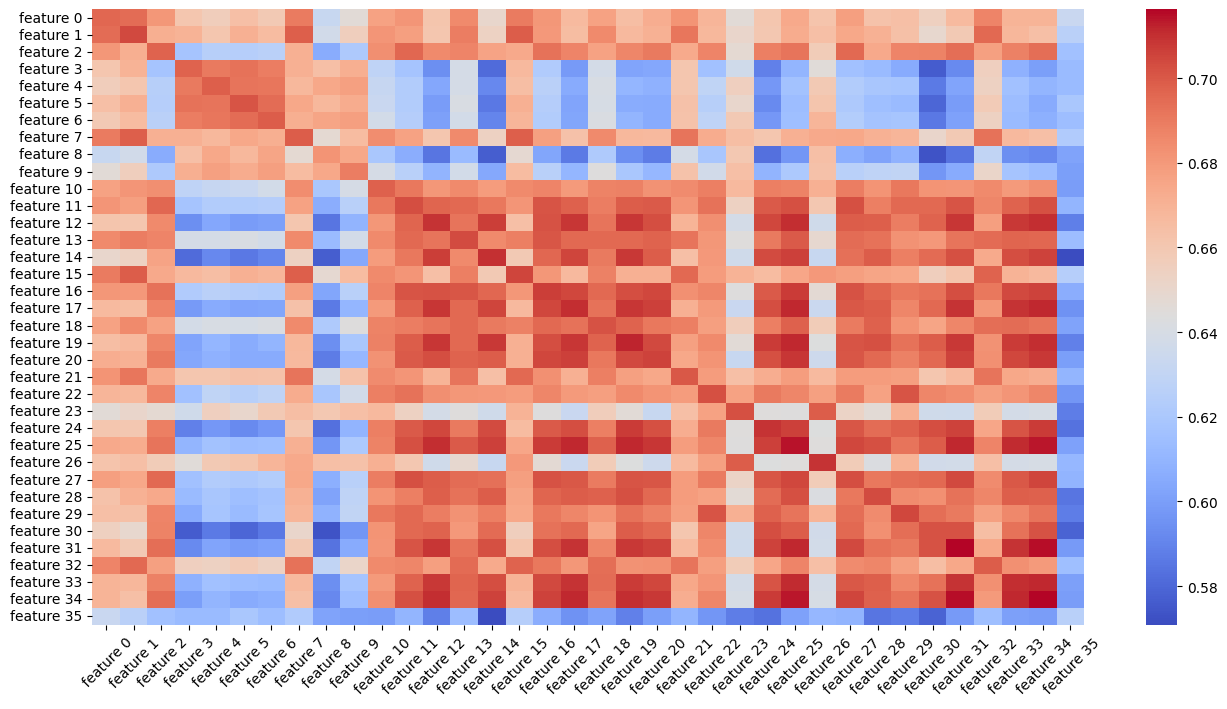}
\end{minipage}
\hfill
\begin{minipage}{0.48\columnwidth}
  \centering
  \includegraphics[width=\linewidth]{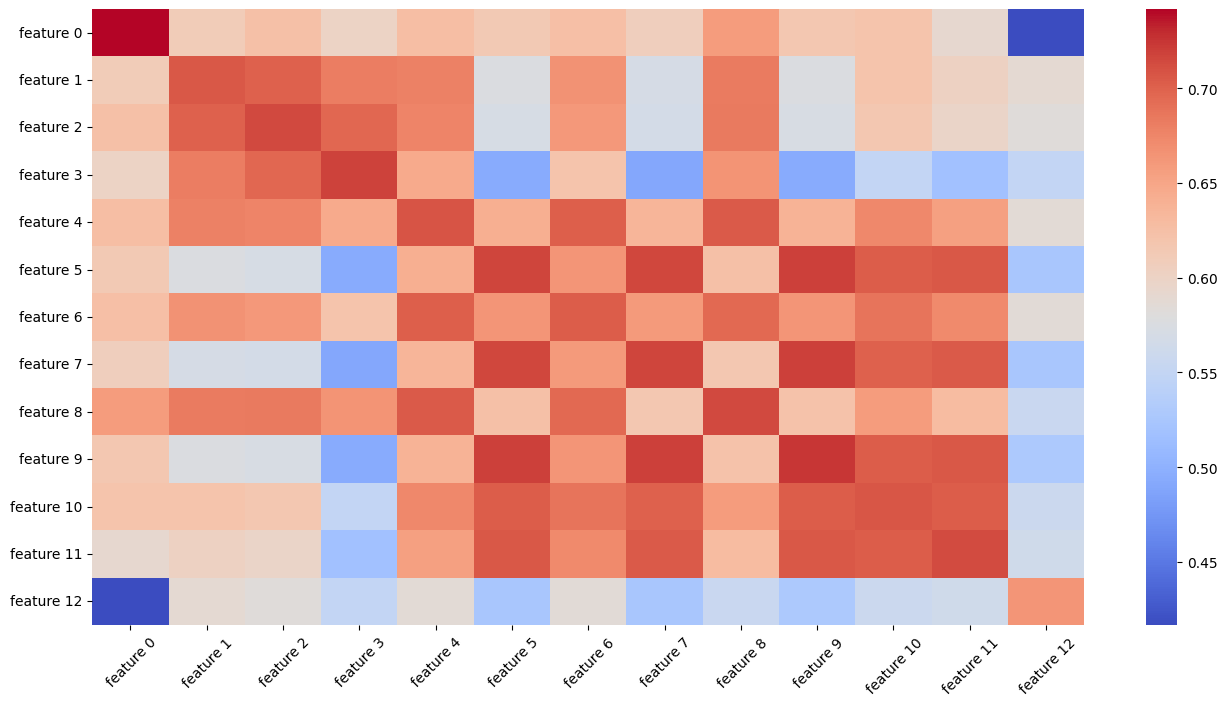}
\end{minipage}

% Row 2
\begin{minipage}{0.48\columnwidth}
  \centering
  \includegraphics[width=\linewidth]{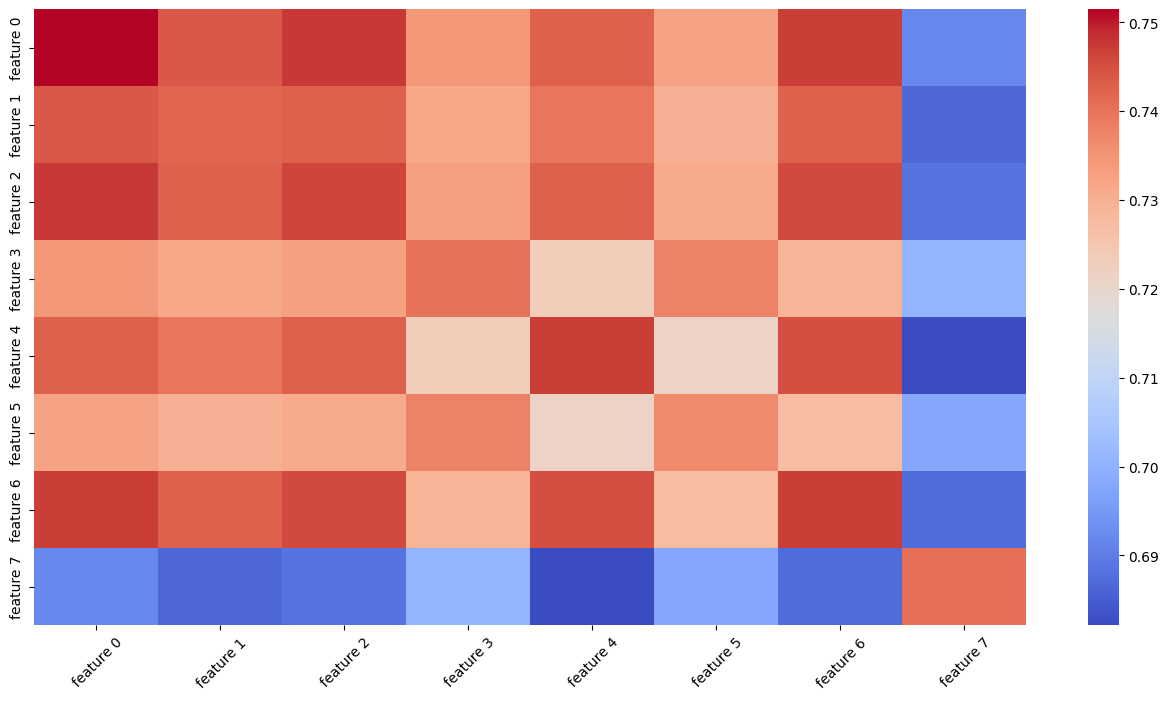}
\end{minipage}
\hfill
\begin{minipage}{0.48\columnwidth}
  \centering
  \includegraphics[width=\linewidth]{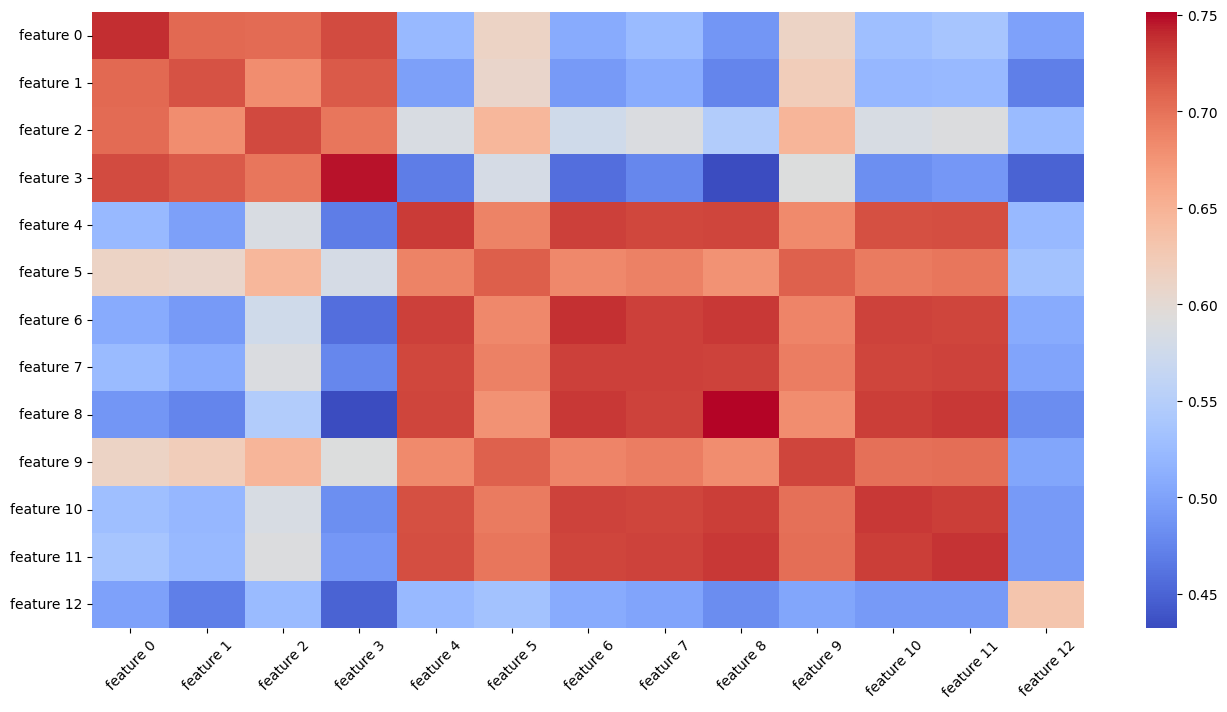}
\end{minipage}

% Row 3
\begin{minipage}{0.48\columnwidth}
  \centering
  \includegraphics[width=\linewidth]{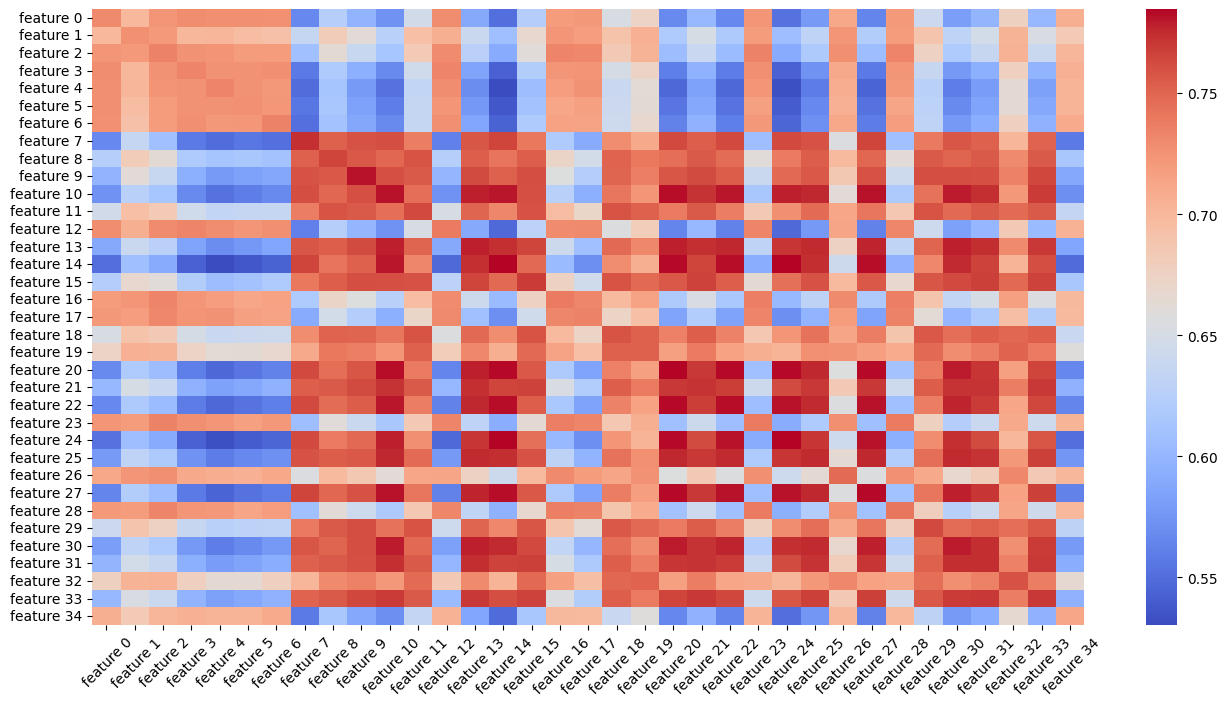}
\end{minipage}
\hfill
\begin{minipage}{0.48\columnwidth}
  \centering
  \includegraphics[width=\linewidth]{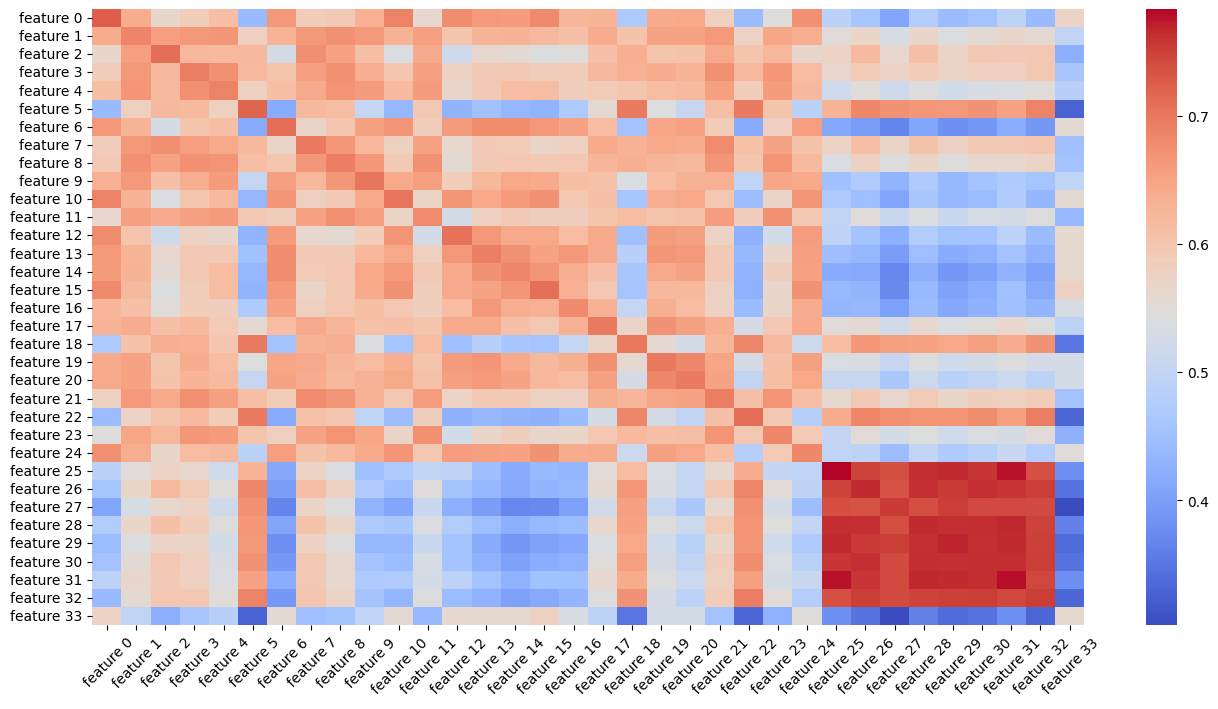}
\end{minipage}

\caption{\textbf{Cosine similarity between multimodal feature embeddings}. Axes denote all tabular and text/image features. From left to right and top to bottom, it shows the correlations between features in the experiments on the PU20, Calc, Cloth, Mass, Petfinder, and Airbnb datasets.}
\label{fig:correlation}
\end{figure}

\paragraph{Cross-Modal Correlation.}
\label{para:correlation}
\Cref{fig:correlation} visualizes cosine similarities among TabPFN–backbone embeddings on all tabular–image and tabular–text datasets. These similarities illustrates the predictive relationships between features learned by MMPFN. As expected, within-modality blocks exhibit high similarity. However, several tabular–image/text pairs are also strongly aligned, indicating that MMPFN models cross-modal interactions rather than only within-modality structure. Details of the cosine similarity computation are provided in supplementary material.

\begin{table}[!t]
\scriptsize
\centering
\caption{\textbf{Performance in Low-data Regime}. Each method uses two rows: accuracy (top) and percentage change vs. full-data (bottom). The ‘Avg.’ column averages percentage changes only. Best results are highlighted in \textbf{bold}.}
\setlength{\tabcolsep}{7.3pt}
\renewcommand{\arraystretch}{1.15}
\begin{tabular}{l|cccc|c}
\toprule
 & PU20 & Mass & Calc & PetFinder & Avg. \\
\midrule
TIP~\cite{du2024tip} 10\%   & 70.44 & 68.31 & 62.27 & 34.86 & 58.97 \\
           & (-10.6) & (-6.58) & (-8.37) & (-6.49) & (-8.00) \\
\addlinespace[2pt]
\midrule
MMPFN 10\% & \textbf{72.87} & \textbf{76.13} & \textbf{72.09} & \textbf{35.73} & \textbf{64.21} \\
           & (-14.14) & (+0.75) & (-5.23) & (-12.30) & (-10.27) \\
\bottomrule
\end{tabular}
\label{tab:small_data}
\end{table}

\paragraph{Robustness in Low-Data Regimes.}
Tabular datasets often require expert annotation, leading to limited sample sizes and sparse labels~\cite{du2024tip, du2025stil}. Models that remain robust under such data scarcity are therefore desirable. In this setting, MMPFN performs strongly. \Cref{tab:small_data} compares MMPFN and TIP when trained on only 10\% of randomly selected samples from each dataset. Although TIP uses self-supervised pretraining on all unlabeled data, we focus on supervised finetuning with limited labeled data.

Although MMPFN shows a larger relative drop, it consistently outperforms TIP on all datasets, even with only 10\% of the data. On CBIS-DDSM Mass, performance even improves under subsampling. This result suggests that the PFN, pretrained on synthetic priors, can better capture discriminative characteristics when finetuned on fewer labeled examples. More details on low-data behavior are provided in the supplementary material.

\begin{figure}[t]
    \centering
    \includegraphics[width=0.85\linewidth]{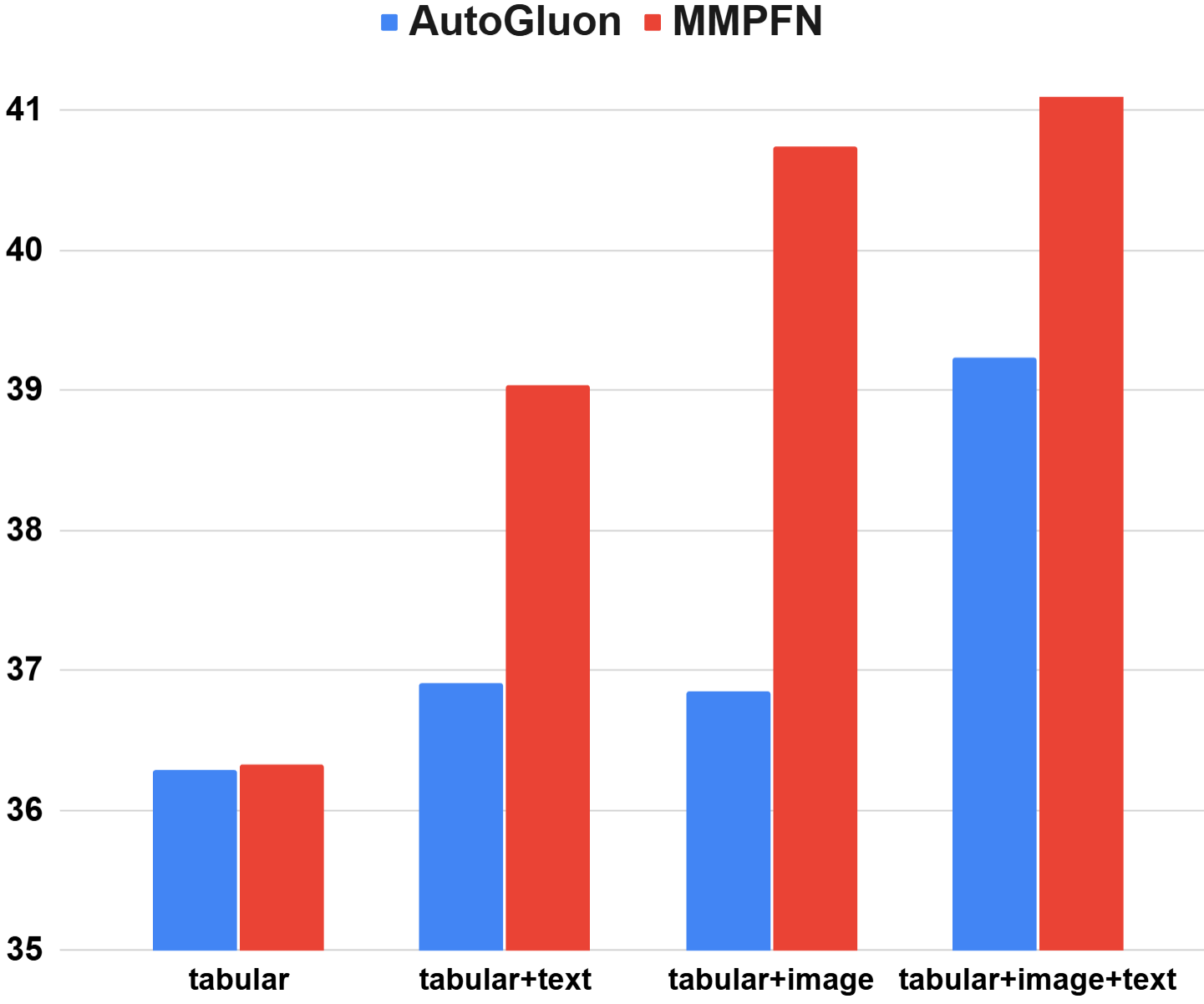} % (a)
    \caption{\textbf{Accuracy of AutoGluon vs.\ MMPFN on PetFinder} under different modality combinations: tabular, +text, +image, +image+text.}
    \label{fig:3modalities}
\end{figure}

\paragraph{Scaling with Added Modalities.}
We assess MMPFN as multiple non-tabular modalities are added. On Petfinder, we compare against AutoGluon, a multimodal AutoML system supporting image and text modalities. As shown in \Cref{fig:3modalities}, MMPFN’s accuracy increases monotonically from \textit{tabular} $\rightarrow$ \textit{tabular+text} $\rightarrow$ \textit{tabular+image} $\rightarrow$ \textit{tabular+image+text} (39\% $\rightarrow$ 40\% $\rightarrow$ 41\%), indicating complementary signal from both image and text. These results have particular significance for tabular modeling, where performance improvements from architectural changes alone are often saturated. Adding complementary modalities offers a practical route to further gains. Moreover, MMPFN outperforms AutoGluon under every combination. Unlike AutoGluon’s large ensembles, MMPFN achieves higher accuracy with a lightweight and specialized architecture.

%% file: sec/5_conclusion.tex
\section{Conclusion}
We introduced MMPFN, a multimodal extension of TabPFN that unifies tabular, image, and text inputs with per-modality encoders, a modality projector, and the TabPFN backbone. We developed MGM and CAP, which map non-tabular embeddings to the tabular space and mitigate token-count–induced attention imbalance. By leveraging pretrained foundation models and fine-tuning lightweight components, MMPFN achieved strong accuracy with substantially lower training costs. Across medical and general-purpose benchmarks, it consistently outperformed competitive state-of-the-art methods, scaled positively as modalities were added, and maintained robust performance in low-data regimes.

%% file: sec/X_suppl.tex
\clearpage
\setcounter{page}{1}
\maketitlesupplementary
\appendix

\setcounter{table}{0}
\setcounter{figure}{0}
\setcounter{section}{0}
\renewcommand{\thetable}{S\arabic{table}}
\renewcommand{\thefigure}{S\arabic{figure}}
\renewcommand{\thesection}{S\arabic{section}}
\renewcommand{\thesubsection}{S\arabic{subsection}}
\renewcommand{\theequation}{S\arabic{equation}}

\section{Additional Analysis}\label{sec:supp_analysis}
\begin{figure*}[!ht]
    \centering
    % -------- Row 1 --------
    \begin{subfigure}{0.48\linewidth}
        \centering
        \includegraphics[width=\linewidth]{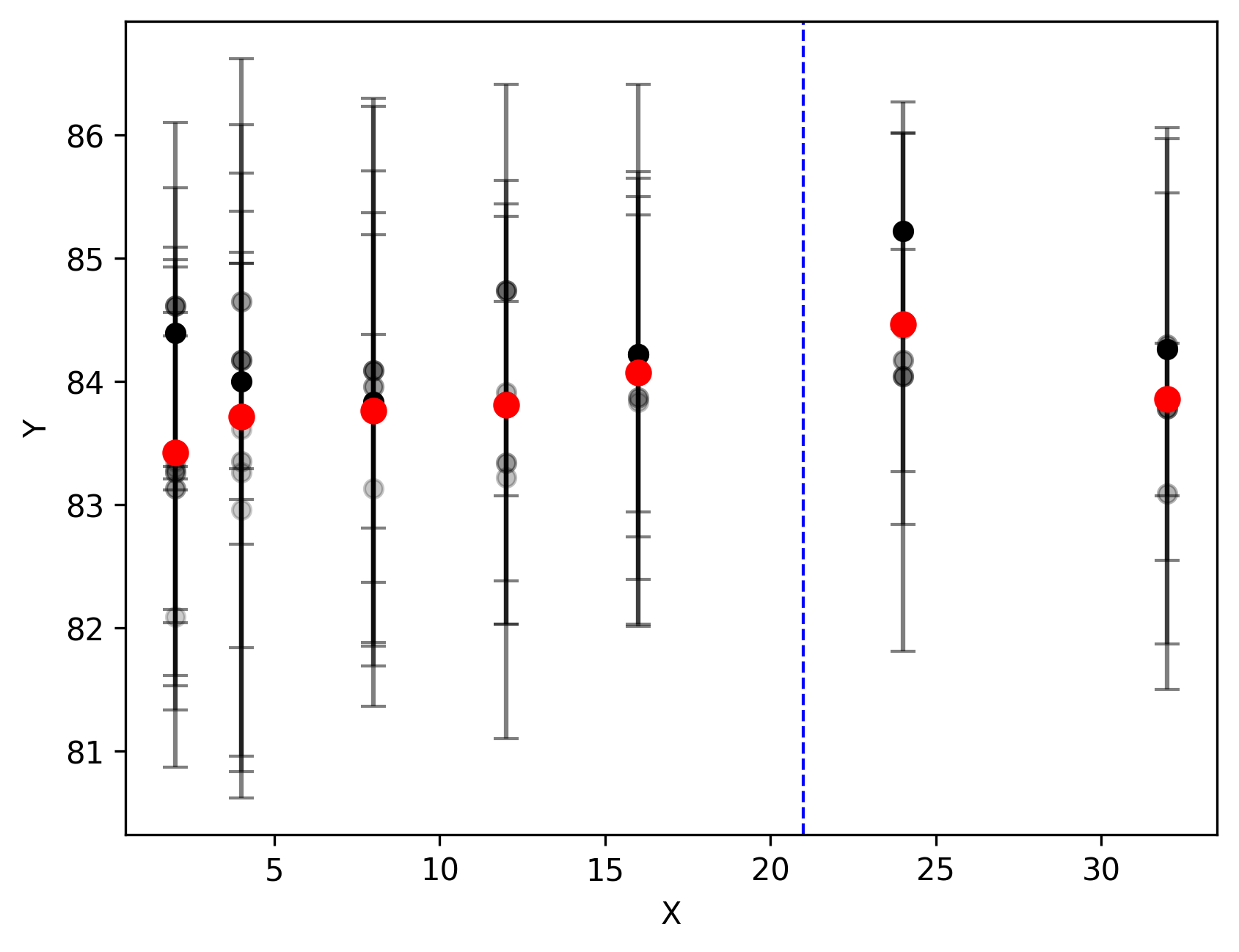}
        \caption{PU20}
    \end{subfigure}
    \hfill
    \begin{subfigure}{0.48\linewidth}
        \centering
        \includegraphics[width=\linewidth]{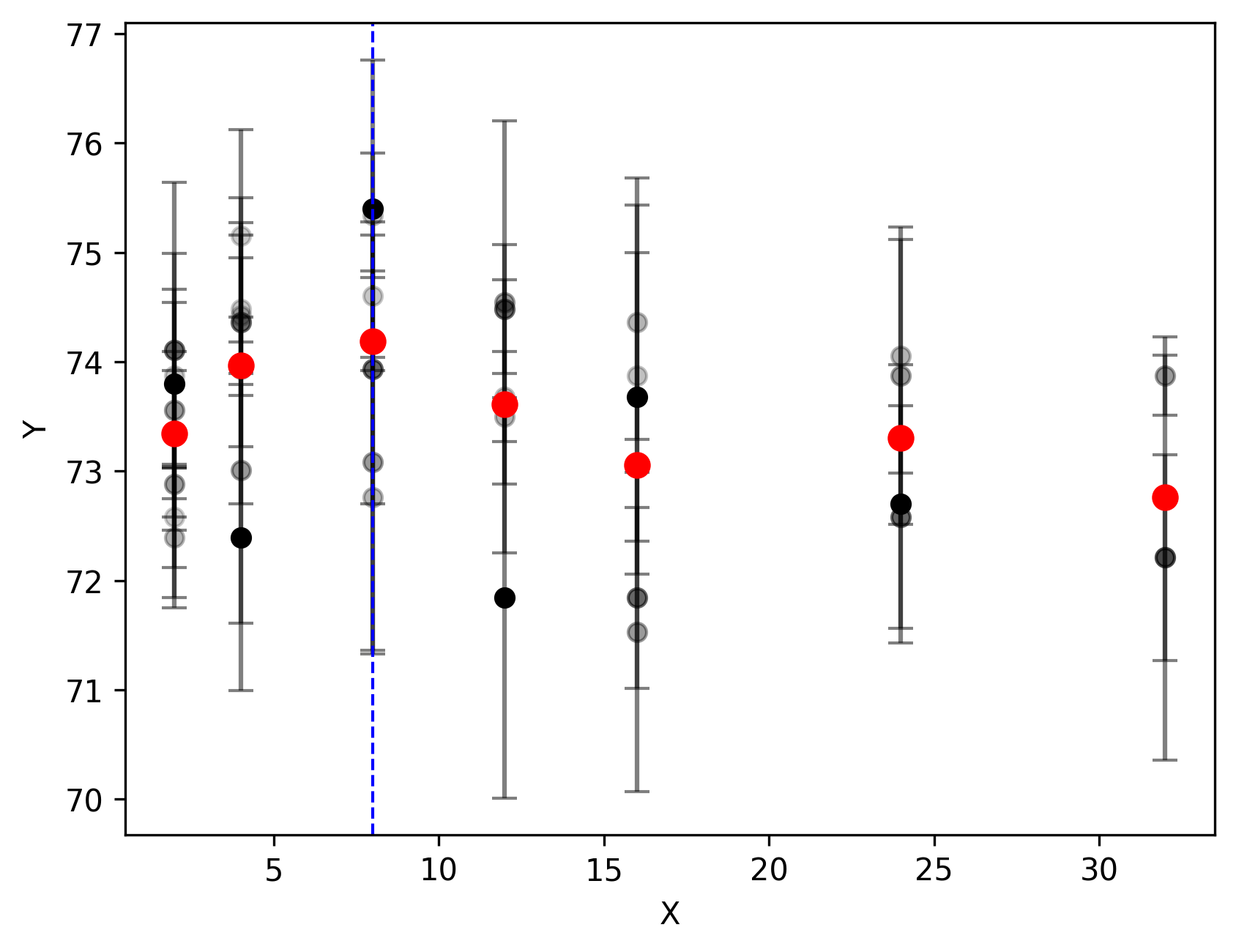}
        \caption{Calc}
    \end{subfigure}
    % -------- Row 2 --------
    \begin{subfigure}{0.48\linewidth}
        \centering
        \includegraphics[width=\linewidth]{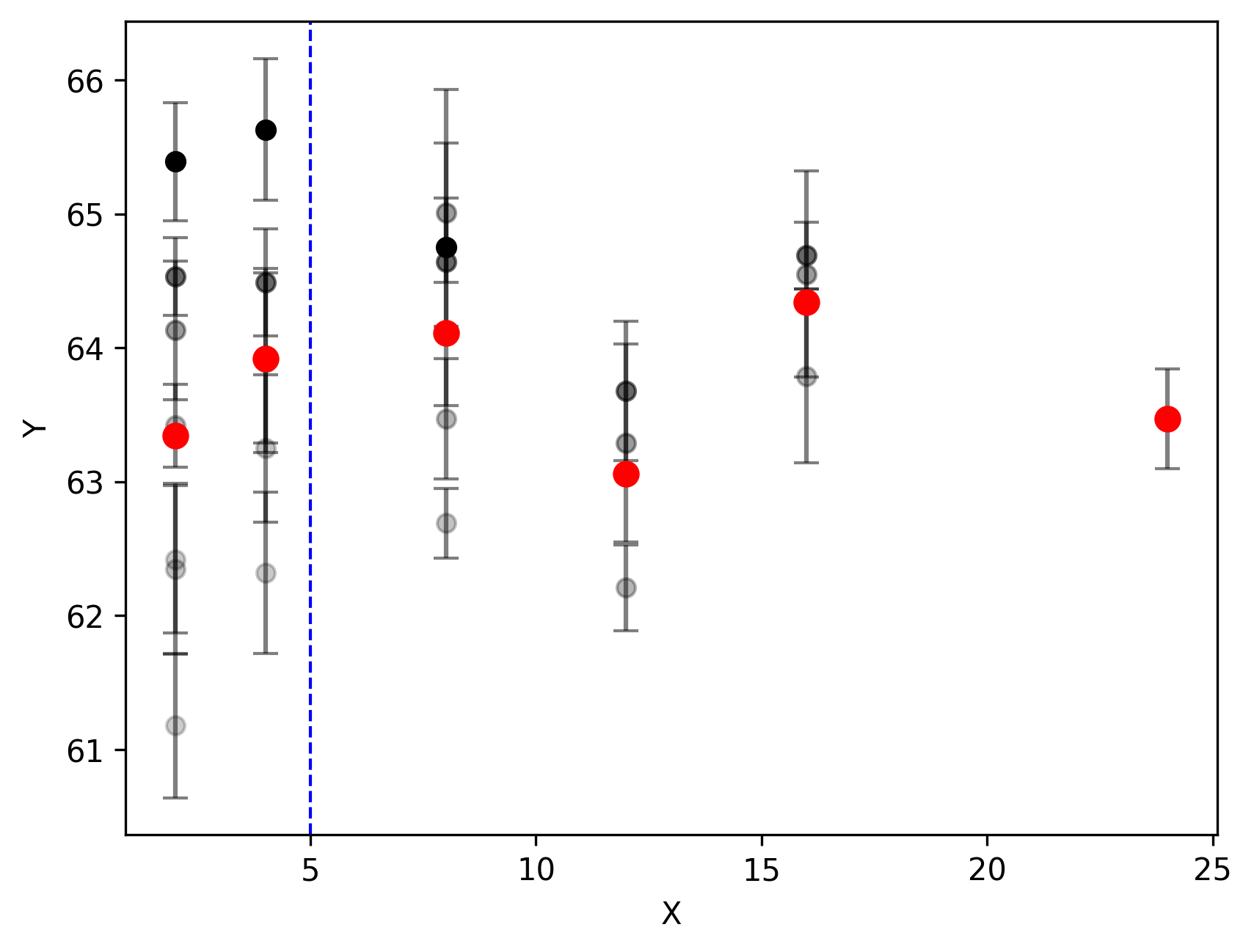}
        \caption{Cloth}
    \end{subfigure}
    \hfill
    \begin{subfigure}{0.48\linewidth}
        \centering
        \includegraphics[width=\linewidth]{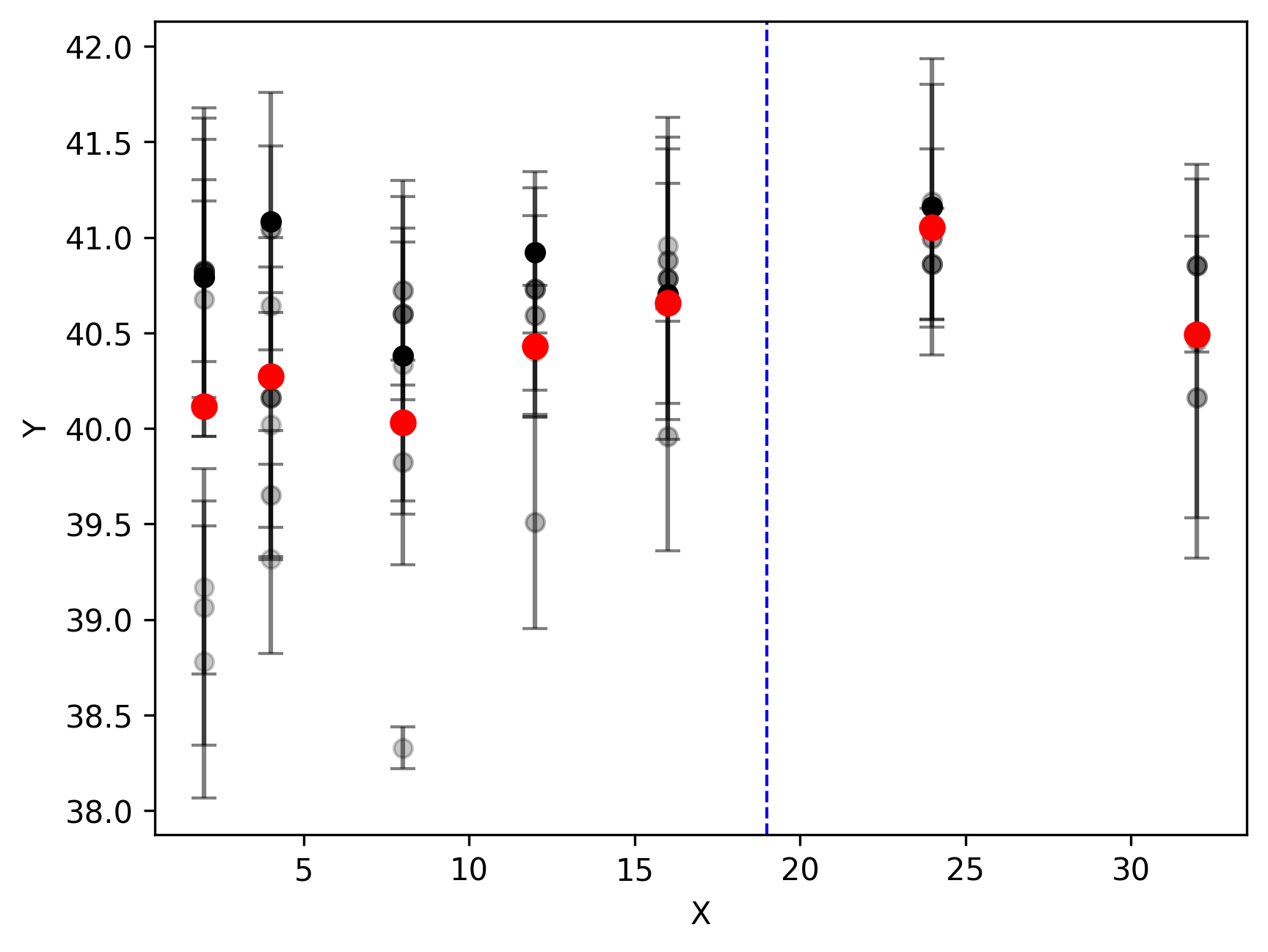}
        \caption{PetFinder-A}
    \end{subfigure}
    \caption{\textbf{Effect of the ratio between tabular and non-tabular features.} The black dots and vertical lines show the mean and variance across five random seeds. Darker black dots correspond to a larger number of MGM heads (i.e., more non-tabular features generated by MGM), ranging from 8 to 128. The red dot indicates the average result across all MGM-head settings. The x-axis shows the number of non-tabular features generated by CAP, and the y-axis denotes accuracy. The blue line represents the number of tabular features. Dataset names are shown above each subfigure.}
    \label{fig:attention_imbalance}
\end{figure*}

\paragraph{Attention Imbalance.}
\Cref{fig:attention_imbalance} illustrates the relationship between the number of non-tabular input features generated by MGM and CAP and the resulting performance across the evaluated datasets. In all cases, the best performance is achieved when the number of non-tabular features is similar to the number of tabular features. 
Performance tends to degrade when the number of non-tabular features is either substantially smaller or larger than the number of tabular features. These results experimentally support our analysis of attention imbalance.

\paragraph{Activation Choice in MGM.}
We study the impact of using GLU as the activation function in MGM on the CBIS--DDSM (MASS) and Salary datasets. \Cref{tab:orthogonality} compares GLU against a GELU baseline in terms of accuracy and mean output-vector orthogonality. Since GLU reduces the dimensionality by half after the gating operation, the GELU baseline is configured with more parameters. Despite this advantage, GLU consistently yields higher accuracy than GELU on both datasets, while also increasing the orthogonality measure. This suggests that the gating mechanism in GLU not only improves predictive performance but also promotes more diverse output representations, aligning with our objective of learning complementary non-tabular features. Consequently, we adopt GLU as the default activation in MGM.

\begin{table}[!ht]
\centering
\scriptsize
\caption{\textbf{Effect of activation choice in MGM.} 
Comparison of accuracy and mean output-vector orthogonality when using GELU versus GLU, with and without orthogonality loss, on CBIS--DDSM (MASS) and Salary.}
{\setlength{\tabcolsep}{7pt}
\begin{tabular}{l|cc|cc}
\toprule
 & \multicolumn{2}{c|}{CBIS--DDSM (MASS)} & \multicolumn{2}{c}{Salary} \\
\midrule
Activation & Accuracy & Orthogonality & Accuracy & Orthogonality \\
\midrule
GELU & 72.09 & 0.0565 & 45.04 & 0.04876 \\
GLU & 75.10 & 0.0913 & 45.87 & 0.05831 \\
\bottomrule
\end{tabular}}
\label{tab:orthogonality}
\end{table}

\paragraph{Parameter budgets in~\Cref{tab:ablation_glu}.} 
In~\Cref{tab:ablation_glu}, the parameter counts of the modality projector vary depending on the architectural choice. Let $N$ denote the number of MGM heads and $d$ the token dimension. The parameter counts scale as follows: single-head + linear $O(d)$, single-head + MLP $O(d^{2}+d)$, multi-head + MLP $O\!\left(N(d^{2}+d)\right)$, and multi-head + MGM $O\!\left(N(d^{2}+d/2)\right)$. MGM requires fewer parameters than the multi-head MLP due to its linear gating with channel splitting. These results indicate that the performance trends in~\Cref{tab:ablation_glu} are not solely explained by increased model capacity. Moreover, the additional parameters introduced by the projector contribute only marginally to inference latency.

\paragraph{Robustness of Low-Data Regimes.}
In~\Cref{tab:small_data}, MMPFN achieves a higher average performance (64.21) than TIP (58.97), despite larger relative drops on PU20 and Petfinder. This robustness primarily stems from strong priors learned during large-scale meta-training on synthetic datasets\cite{hollmann2023tabpfn, hollmann2025accurate}, which capture a broad range of plausible tabular distributions and enable effective generalization from few real samples. 
Fine-tuning then provides a light task-specific adaptation on top of this Bayesian inference. Because the model requires only light adaptation, it avoids overfitting and remains stable in low-sample settings. Together with the inductive bias of the Per-Feature Transformer, this explains the superior performance of our MMPFN across low-data experiments.

\paragraph{Replacing Modality Encoders.}
MMPFN combines pretrained models, making the framework naturally extensible as newer and stronger encoders become available. This modular design allows components such as TabPFN or DINO to be replaced with more recent architectures without altering the overall pipeline. Leveraging improved pretrained models enhances the quality of feature representations and can lead to measurable downstream gains. For example, substituting DINOv2 with the recently released DINOv3 yields consistent improvements across datasets, as shown in~\Cref{tab:v3}; on PU20, accuracy increases by approximately $0.74$ percentage points. 

We also examine the effect of replacing the text encoder. As shown in~\Cref{tab:text_encoders}, switching between ELECTRA and DeBERTa results in only minor performance differences across the evaluated datasets. These results suggest that while stronger encoders can provide modest improvements, the overall performance of MMPFN remains relatively stable across different encoder choices.

% Similarly, TabPFNv2\cite{hollmann2025accurate} outperforms TabPFNv1\cite{hollmann2023tabpfn} across all evaluated tasks. Accordingly, MMPFN adopts TabPFNv2 as the default tabular encoder in all experiments

\begin{table}[!ht]
\centering
\scriptsize
\caption{\textbf{Effect of replacing the image encoder.} Performance of MMPFN when substituting DINOv2 with ResNet50 and DINOv3. Results are reported as averaged accuracy over five random seeds.}
{
\setlength{\tabcolsep}{10pt}
\begin{tabular}{l|cccc}
\toprule
Encoder & PU20 & Mass & Calc & Petfinder \\
\midrule
ResNet50 & 83.26 &   -   & 73.94 &   -   \\
DINOv2   & 85.22 & 74.53 & 75.40 & 40.74 \\
DINOv3   & 85.61 & 75.48 & 76.75 & 40.57 \\
\bottomrule
\end{tabular}
}
\label{tab:v3}
\end{table}

\begin{table}[!ht]
\centering
\scriptsize
\caption{\textbf{Effect of replacing the text encoder.} Performance of MMPFN when using different pretrained text encoders. Results are reported as averaged accuracy over five random seeds.}
{
\setlength{\tabcolsep}{15pt}
\begin{tabular}{l|cc}
\toprule
Encoder & Airbnb & Salary \\
\midrule
Electra  & 47.78 & 46.17 \\
DeBERTa  & 47.82 & 45.69 \\
\bottomrule
\end{tabular}
}
\label{tab:text_encoders}
\end{table}

\begin{figure*}[!t]
\centering
\includegraphics[width=\linewidth]{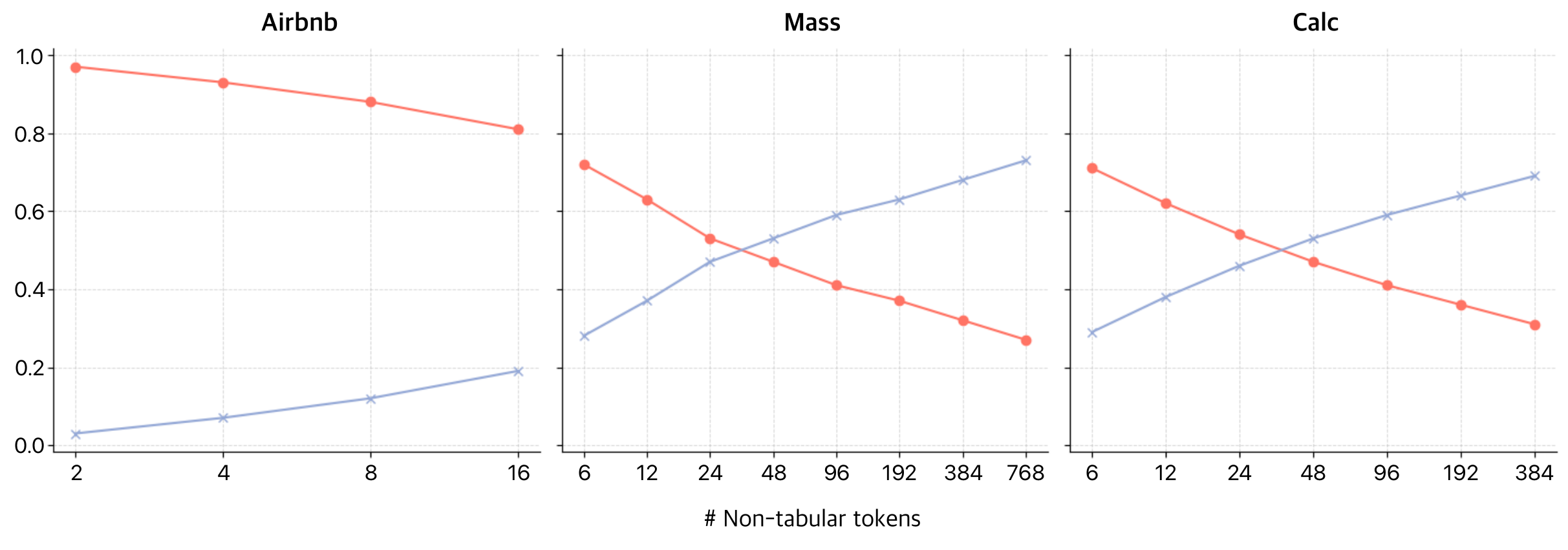}
\caption{\textbf{Additional results of token-count and attention mass}. We extend the analysis in \Cref{fig:token_attention} to additional datasets. For Mass and Calc, each sample contains three images, and thus the number of non-tabular tokens is given by the number of mgm heads times 3. The number of tabular tokens is 25, 4 and 4 for Airbnb, Mass, and Calc, respectively.}
\label{fig:token_attention_sup}
\end{figure*}

\begin{figure}[!ht]
    \centering
    \begin{subfigure}{\linewidth}
        \centering
        \includegraphics[width=\linewidth]{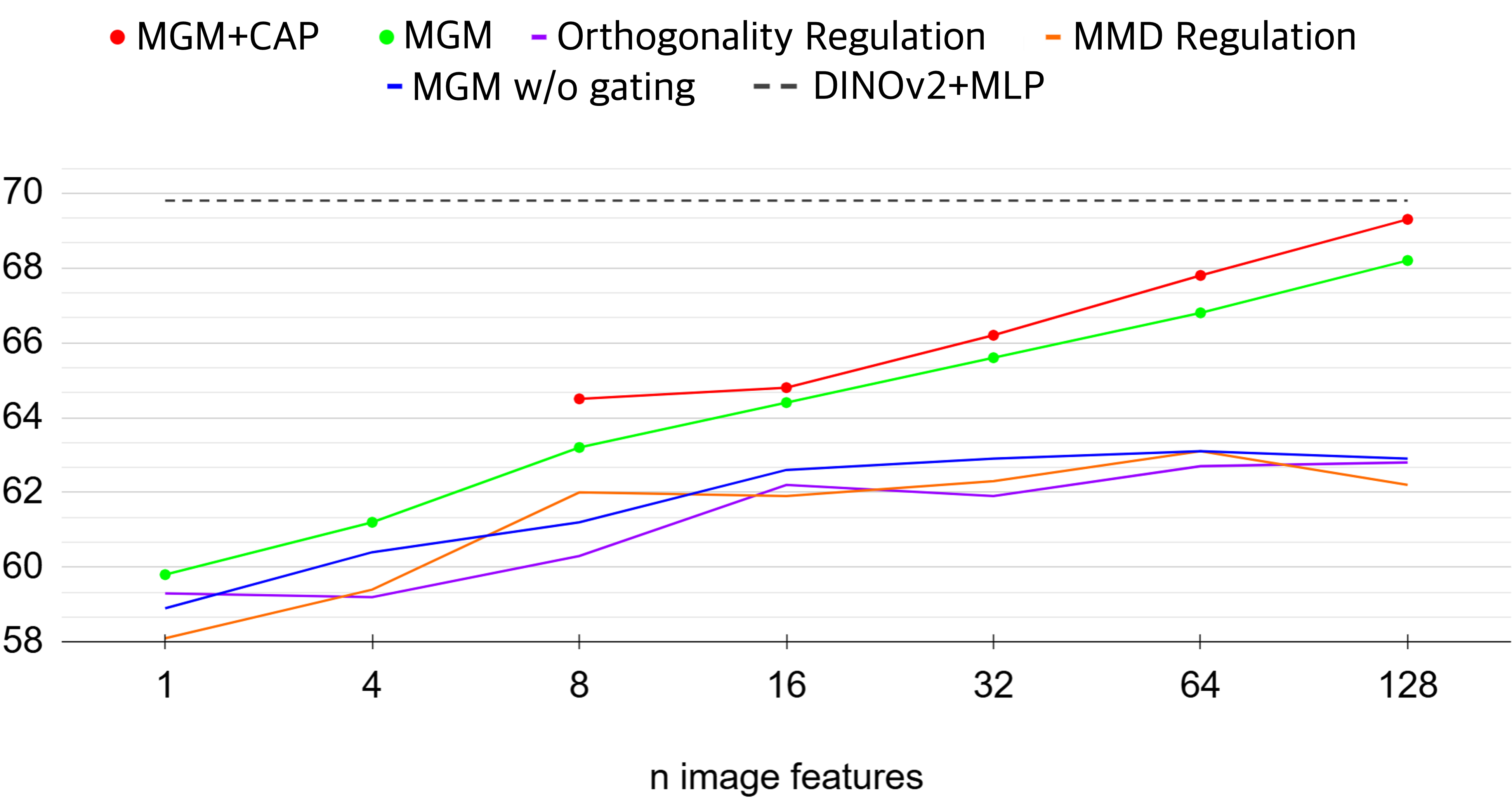}
        \caption{Image-only MGM variants.}
    \end{subfigure}
    \begin{subfigure}{\linewidth}
        \centering
        \includegraphics[width=\linewidth]{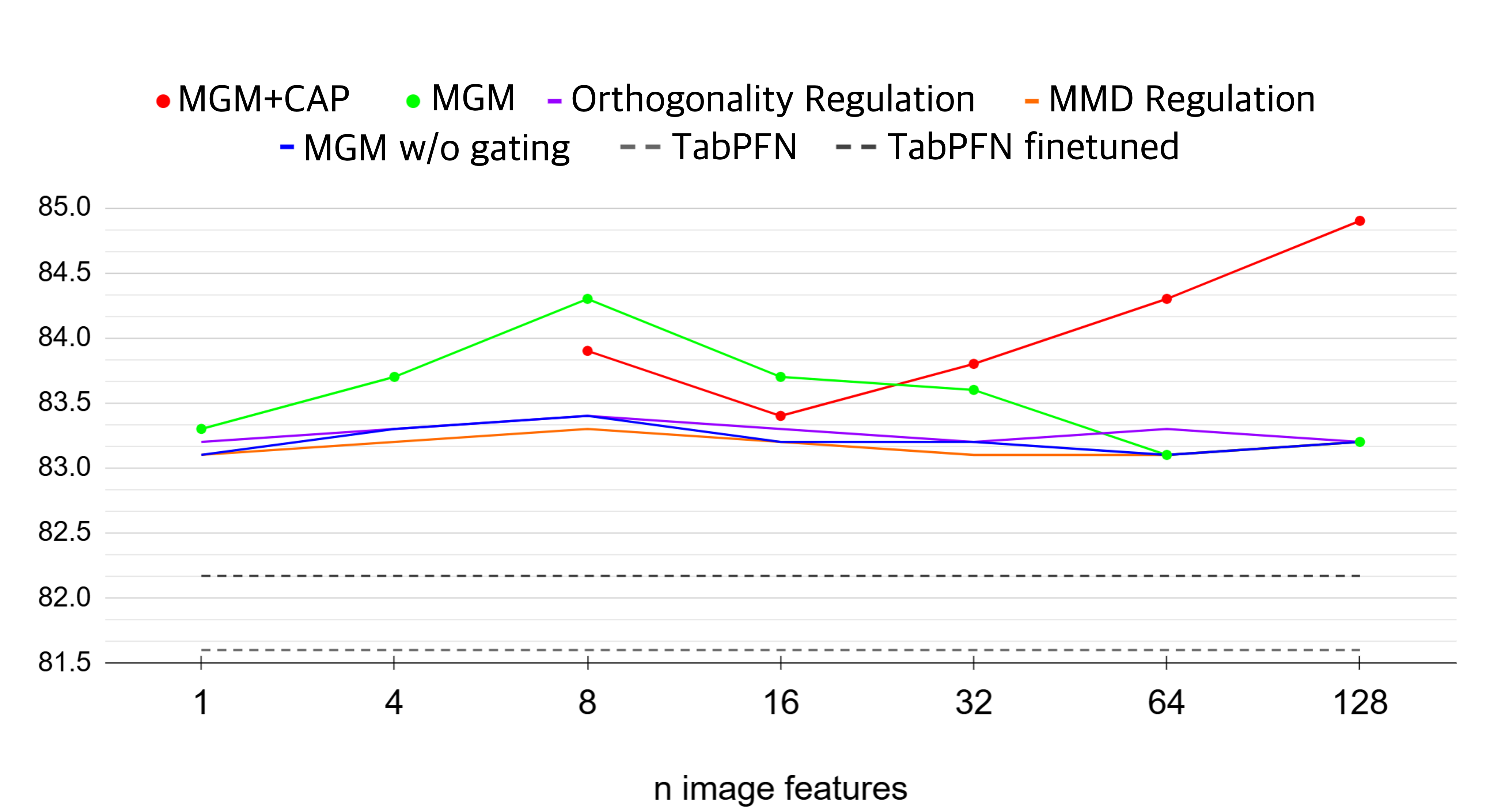}
        \caption{Full MMPFN (tabular + image) variants.}
    \end{subfigure}
    \caption{\textbf{Effect of distribution-alignment and orthogonality regularization.}
    Under the same experimental conditions as~\Cref{fig:perf_by_n_nontab_pu20}, we compare the performance of variants that incorporate orthogonality and MMD-based regularization constraints to the MGM baseline, for both image-only and tabular+image settings.}
    \label{fig:perf_with_reg}
\end{figure}

\paragraph{Distribution Alignment Across Modalities.}
We investigate whether applying embedding-space alignment techniques—commonly used in multimodal learning—can further improve the performance of MMPFN. In particular, we incorporate Maximum Mean Discrepancy (MMD)~\cite{gretton2012kernel}, a standard measure of distributional discrepancy~\cite{zhu2024cross}, into our framework. We apply MMD to the embeddings generated for each feature to reduce the distributional gap between representations from different modalities. For tabular data, where feature distributions can differ substantially across dimensions, we additionally employ Joint MMD (JMMD)~\cite{long2017deep} to capture discrepancies at the level of the joint feature distribution.

We train the multi-head MLP module that produces unstructured feature embeddings by adding the discrepancy between tabular embeddings and image/text embeddings as an auxiliary loss term. However, as shown in~\Cref{fig:perf_with_reg}, this alignment-based regularization consistently underperforms the MGM baseline, and incorporating the same loss directly into MGM also fails to yield improvements. Together with our cosine-similarity analysis, these negative results suggest that embeddings extracted by MGM from image and text modalities are already mapped into a semantically compatible space with tabular embeddings, enabling effective interaction through the attention module. Moreover, while MMD-style losses can reduce distributional gaps, they may also suppress discriminative variations, leading to performance degradation. We therefore infer that once MGM and CAP produce sufficiently aligned embeddings, enforcing additional distributional alignment does not provide further gains and can even be detrimental.

\paragraph{Comparison with Patch-Token Features.}
In MMPFN, the text encoder~\cite{clark2020electra} and the image encoder~\cite{oquabdinov2} produce output embeddings for every token (e.g., text tokens or image patches). In principle, one could replace the $[CLS]$-based MGM features with ViT patch-token outputs and use all token embeddings directly for feature generation. However, this design has both practical and empirical drawbacks. From a memory perspective, using all patch tokens is substantially more expensive than using the aggregated $[CLS]$ token. For example, when resizing PAD-UFES-20 images to $336 = 14 \times 24$ pixels and encoding them with the DINOv2 ViT-B/14 backbone, the model produces $576$ token embeddings per image—more than four times the number of MGM heads ($128$) used in our experiments. In terms of storage, the $[CLS]$ embedding requires only $7.1$ MB, whereas retaining all patch-token outputs occupies $4.1$ GB, leading to a prohibitive increase in memory consumption. A similar issue arises for text: in the Cloth dataset, many text attributes approach the maximum input length of $512$ tokens, so storing all token embeddings again results in excessive memory usage.

Empirically, we also observe that models using ViT patch-token outputs underperform those relying on the $[CLS]$-driven MGM features. On PU20, replacing the $[CLS]$-based MGM heads with patch-token features decreases accuracy by $0.85$ percentage points (from $85.22\%$ to $84.02\%$). We attribute this degradation to the fact that the $[CLS]$ representation of a well-trained foundation model already encodes task-relevant global information, while raw token-level outputs contain substantial redundancy and noise. Consequently, patch-token features are less suitable than $[CLS]$-based MGM heads for constructing compact, tabular-like feature representations.

\paragraph{Additional qualitative results of attention imbalance.}
To complement the analysis in the main paper, we further examine the relationship between token count and attention mass on additional datasets. As shown in \Cref{fig:token_attention_sup}, we observe a consistent trend across all datasets: as the number of non-tabular tokens increases, the attention mass assigned to the non-tabular modality increases monotonically, while the attention to tabular tokens decreases accordingly.

For datasets such as Mass and Calc, each sample contains multiple images, and the number of non-tabular tokens is given by the number of MGM heads multiplied by the number of images. Despite variations in the number of tabular tokens across datasets, such as 25 for Airbnb and 4 for Mass and Calc, the same qualitative behavior consistently emerges. These results further indicate that attention allocation is primarily determined by the relative proportion of tokens within the input sequence.

\begin{figure}[!ht]
    \centering
    \begin{subfigure}{0.48\linewidth}
        \centering
        \includegraphics[width=\linewidth]{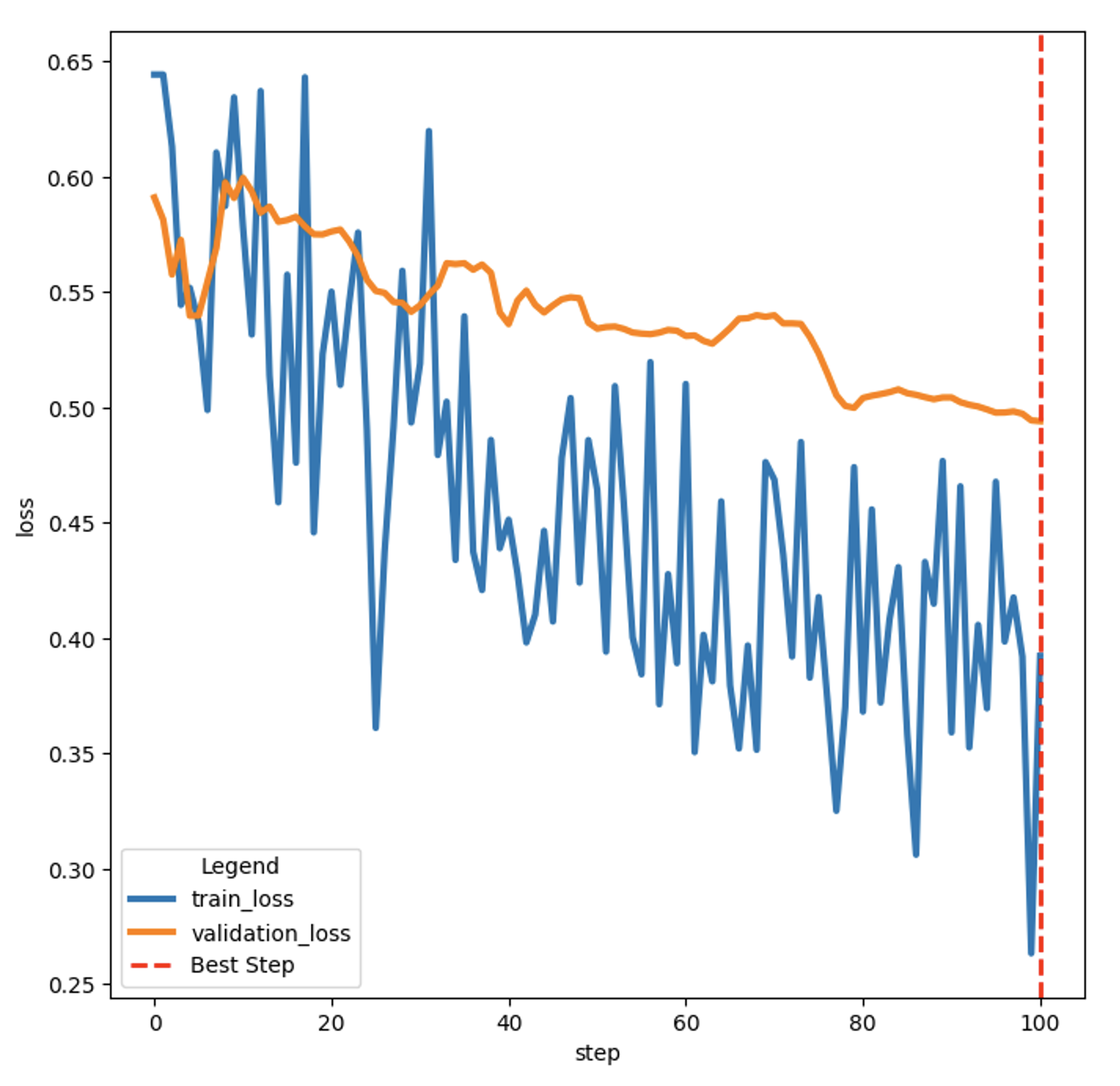}
        \caption{PU20}
    \end{subfigure}
    \hfill
    \begin{subfigure}{0.48\linewidth}
        \centering
        \includegraphics[width=\linewidth]{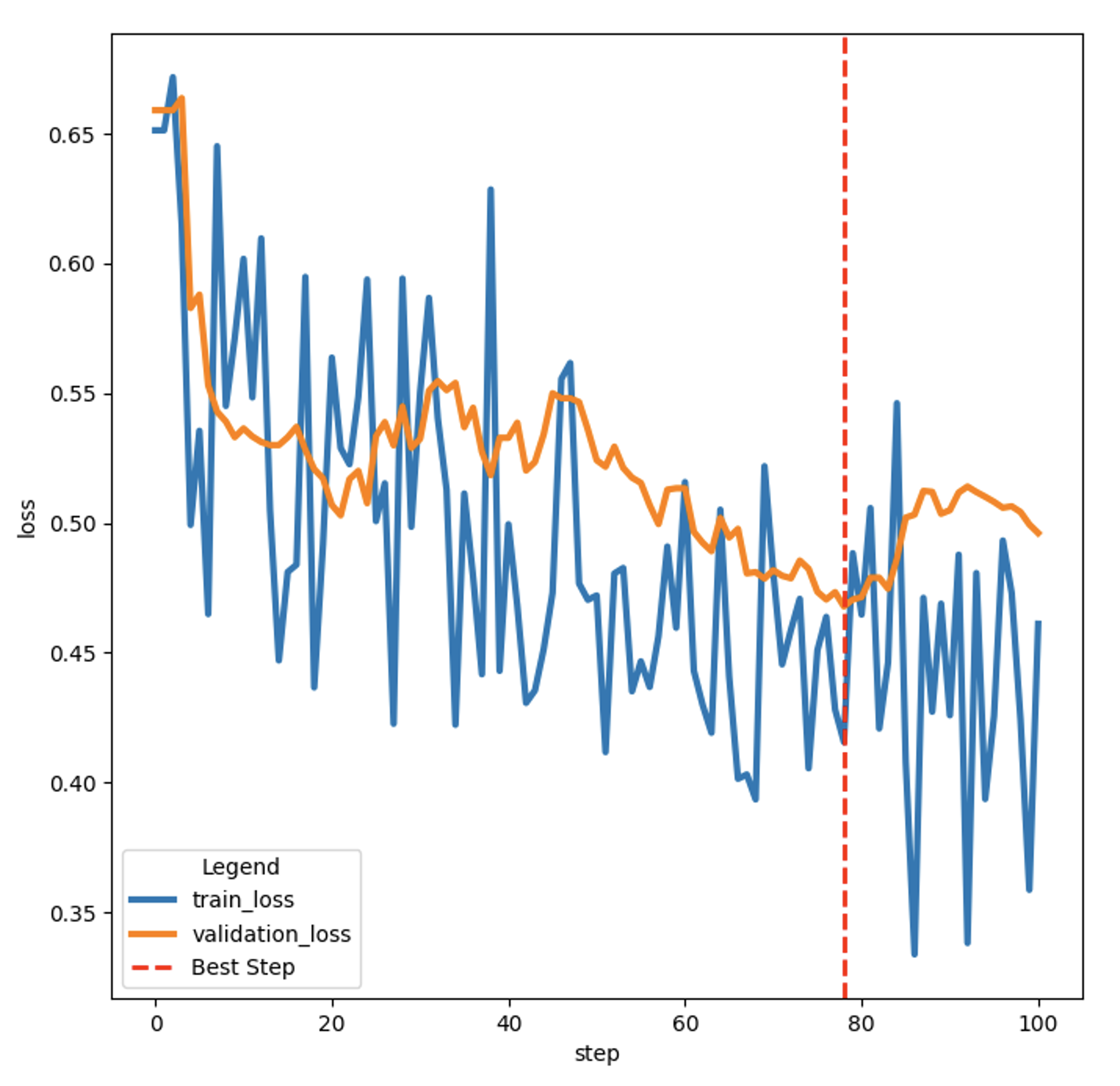}
        \caption{Mass}
    \end{subfigure}
    \begin{subfigure}{0.495\linewidth}
        \centering
        \includegraphics[width=\linewidth]{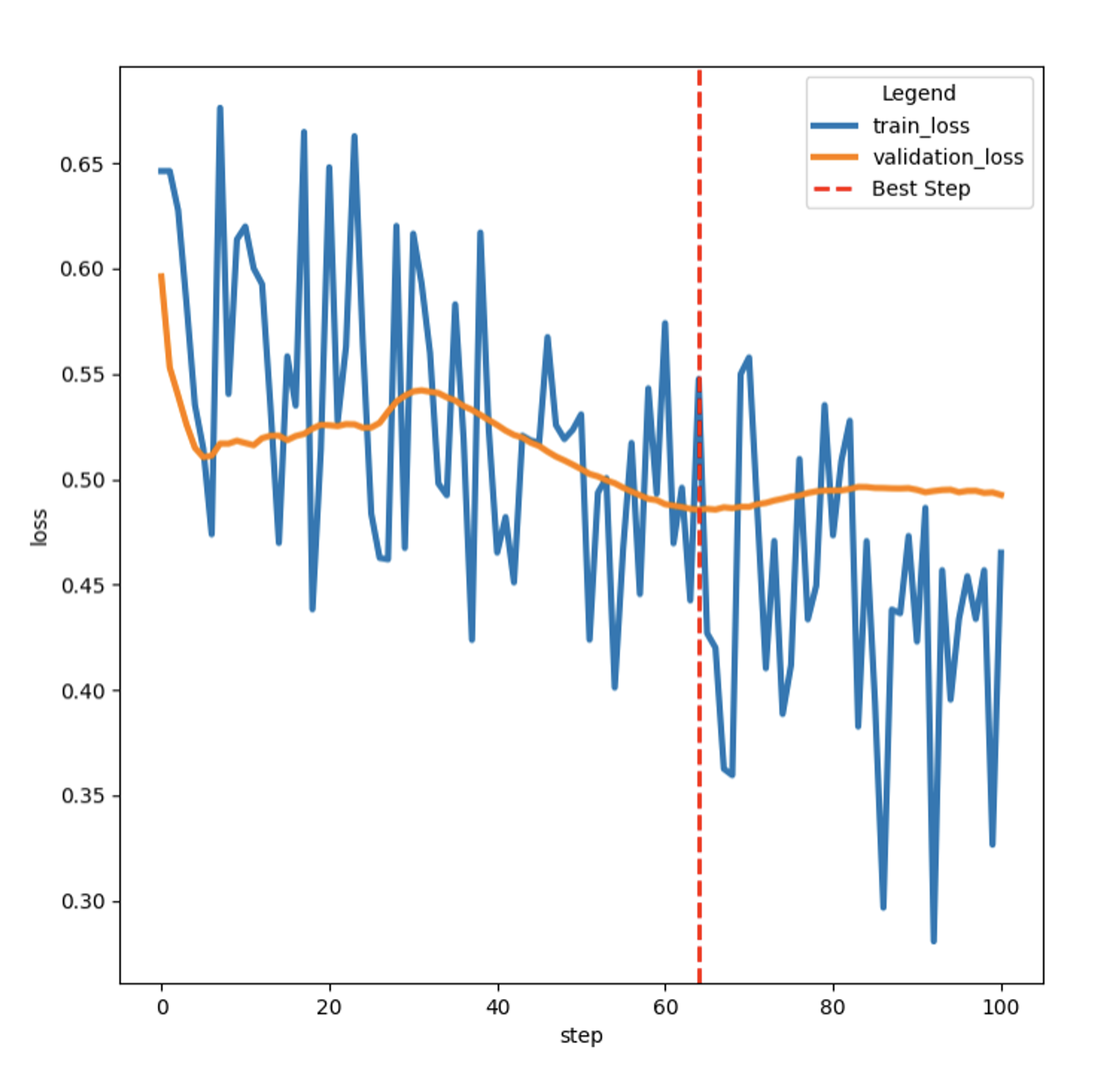}
        \caption{Calc}
    \end{subfigure}
    \hfill
    \begin{subfigure}{0.495\linewidth}
        \centering
        \includegraphics[width=\linewidth]{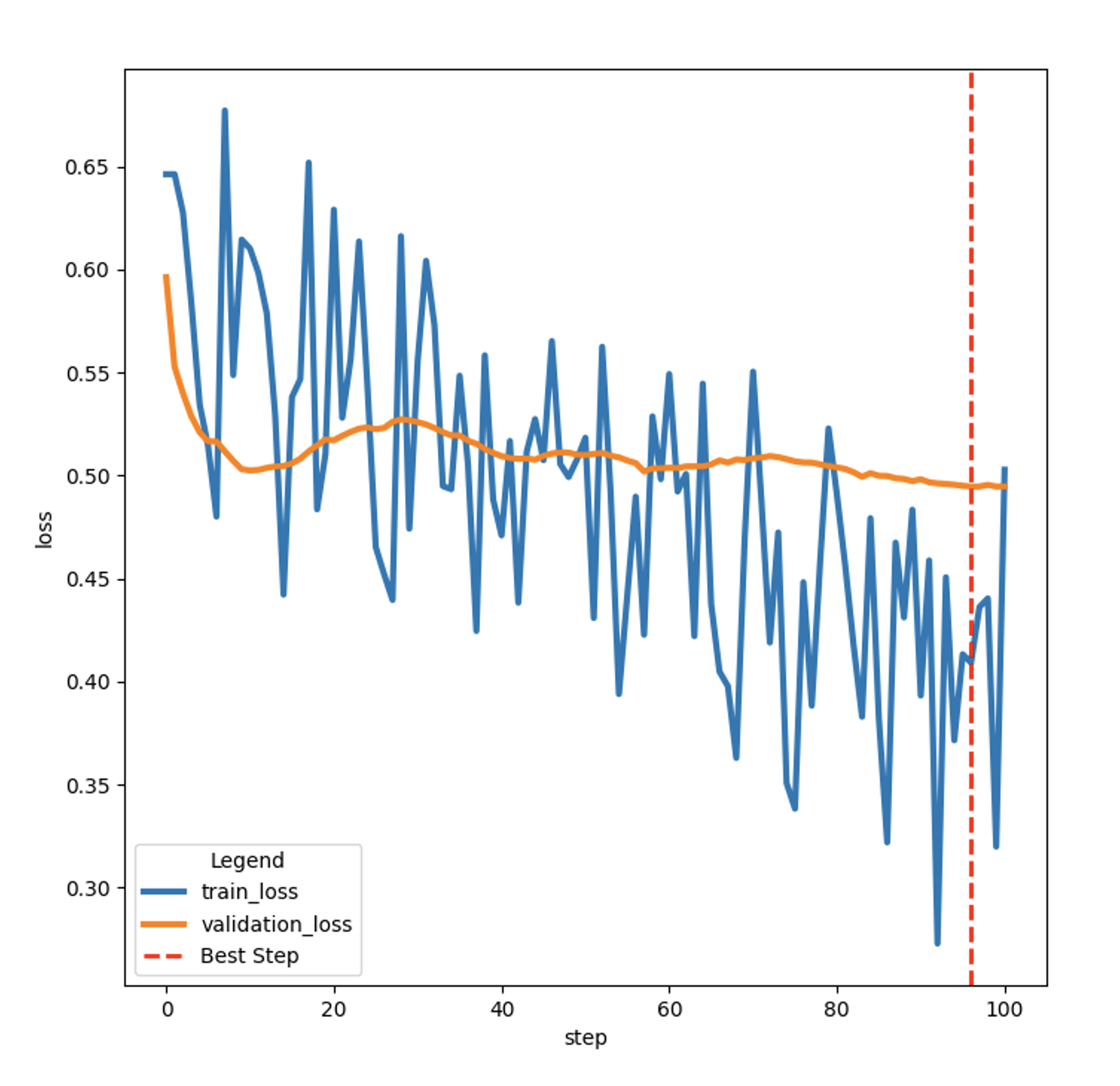}
        \caption{PetFinder-A}
    \end{subfigure}
    \caption{\textbf{Fine-tuning loss curves of MMPFN across datasets.}
    Each subfigure shows the validation loss over training steps for four different datasets, illustrating stable convergence behavior under our fine-tuning setup.}
    \label{fig:loss_curves}
\end{figure}

\paragraph{Implementation Details.}
For the training procedure, we adopt the official TabPFN repository\footnote{\url{https://github.com/LennartPurucker/finetune_tabpfn_v2}} and modify it to support MMPFN by extending the \texttt{MMPFNClassifier} and \texttt{MMPFNRegressor} classes. Fine-tuning is performed by splitting the available data into training and validation sets and updating model parameters based on the validation loss. We use a small learning rate of $1\times10^{-5}$, a fixed budget of $100$ training steps, and ScheduleFree~\cite{defazio2024road} for learning rate scheduling. \Cref{fig:loss_curves} shows the learning curve on four different datasets. For contrastive-pretraining baselines~\cite{hager2023best,du2024tip,du2025stil}, we train for $500$ epochs using a cosine-annealing scheduler with a $10$-epoch warmup.

\paragraph{Text Data Pre-Processing.}
We adopt the text pre-processing pipeline of TTT~\cite{bonnier2024revisiting}, following the implementations provided in the official codebase. For the Salary dataset, the original source URL referenced in \citet{bonnier2024revisiting} is no longer accessible, so we use a Kaggle-hosted copy instead. Applying the official TTT scripts to this version does not reproduce the exact dataset size reported in the paper, suggesting minor discrepancies. We therefore re-evaluate the TTT baseline on this revised dataset; the resulting accuracy (46.5) closely matches the originally reported performance, confirming that the new version is suitable for evaluating our model.

Both ELECTRA and DeBERTa text encoders are limited to 512 input tokens, so longer sequences are truncated. For datasets with multiple text attributes, we extract embeddings for each attribute separately and incorporate them as additional text features, whereas TTT concatenates all text columns into a single sequence; this yields small but consistent accuracy gains, although the improvements remain within the error margin and are not reported in the main comparison table. The Airbnb and PetFinder datasets contain Chinese characters in their text fields; because the ELECTRA variant we use is not pretrained on Chinese, these characters are replaced with empty strings before encoding. For CatBoost and AutoGluon, we rely on the libraries’ built-in text handling capabilities.

\paragraph{Cosine Similarity Computation.}
We provide the detailed procedure used for the cosine-similarity–based correlation analysis in~\Cref{fig:correlation,para:correlation}. The TabPFN encoder for tabular data normally groups multiple features into a single embedding to reduce memory usage, but such grouping can introduce noise when comparing cosine similarity between tabular and image (or text) embeddings. To obtain more precise relationships, we set the tabular group size to $1$, generate an individual embedding for each feature, and then compare these feature-wise embeddings with image embeddings.

Cosine similarity between input features is computed at the instance level and subsequently averaged across instances. Averaging embeddings over the entire dataset before computing similarity vectors would be cheaper but tends to underrepresent the contribution of individual samples, whereas computing similarity for every token embedding is prohibitively expensive and makes global patterns difficult to visualize. We therefore adopt an intermediate strategy, which balances computational cost and fidelity.